%% file: iclr2023_conference.tex
\pdfoutput=1 
\documentclass{article} 
\usepackage{iclr2023_conference,times}

\input{math_commands.tex}

\usepackage[hidelinks]{hyperref}
\usepackage{url}
\usepackage{times}
\usepackage{latexsym}
\usepackage{graphicx}
\usepackage{float}
\usepackage{subfloat}
\usepackage{caption}
\usepackage{truncate}
\usepackage{ragged2e}
\usepackage{inputenc}
\usepackage{csquotes}
\usepackage{booktabs}
\usepackage{multirow}
\usepackage{amsmath}
\usepackage{amssymb}
\usepackage{pifont}
\usepackage{xurl}
\usepackage{xinttools}
\usepackage{ifthen} 
\usepackage{todonotes}
\usepackage[T1]{fontenc}
\usepackage{enumitem}
\usepackage{tabularx}
\usepackage{tabulary}
\newcommand{\cmmnt}[1]{}
\newcommand{\cmark}{\ding{51}}%
\newcommand{\xmark}{\ding{55}}%
\graphicspath{{resources/}}

\title{Bidirectional Language Models Are Also Few-shot Learners}

\author{Ajay Patel \thanks{ Correspondence to: \texttt{ajayp@upenn.edu}} \\
University of Pennsylvania\\
\And
Bryan Li \\
University of Pennsylvania \\
\And
Mohammad Sadegh Rasooli \\
Microsoft \\
\AND
Noah Constant \\
Google Research \\
\And
Colin Raffel \\
UNC Chapel Hill \\
\And
Chris Callison-Burch \\
University of Pennsylvania \\
}

\iclrfinalcopy 
\begin{document}

\input{./resources/figures}
\input{./resources/few-shot_results}
\input{./resources/zero-shot_results}
\input{./resources/ablation_tables}
\input{./resources/xquad_results}
\input{./resources/squad_and_summarization_results}
\input{./resources/selected_generations}

\maketitle

\begin{abstract}
Large language models such as GPT-3 \citep{gpt3} can perform arbitrary tasks without undergoing fine-tuning after being prompted with only a few labeled examples. An arbitrary task can be reformulated as a natural language prompt, and a language model can be asked to generate the completion, indirectly performing the task in a paradigm known as prompt-based learning. To date, emergent prompt-based learning capabilities have mainly been demonstrated for unidirectional language models. However, bidirectional language models pre-trained on denoising objectives such as masked language modeling produce stronger learned representations for transfer learning. This motivates the possibility of prompting bidirectional models, but their pre-training objectives have made them largely incompatible with the existing prompting paradigm. We present \textsc{Sap}  (Sequential Autoregressive Prompting), a technique that enables the prompting of bidirectional models. Utilizing the machine translation task as a case study, we prompt the bidirectional mT5 model \citep{xue-etal-2021-mt5} with \textsc{Sap} and demonstrate its few-shot and zero-shot translations outperform the few-shot translations of unidirectional models like GPT-3 and XGLM \citep{xglm}, despite mT5's approximately 50\% fewer parameters. We further show \textsc{Sap} is effective on question answering and summarization. For the first time, our results demonstrate prompt-based learning is an emergent property of a broader class of language models, rather than only unidirectional models.
\end{abstract}

\section{Introduction}
\label{sec:introduction}

Recent work on GPT-2~\citep{gpt2} and GPT-3~\citep{gpt3} have shown that large language models possess few-shot learning capabilities and zero-shot instruction following capabilities, despite only being pre-trained with a self-supervised causal language modeling objective (which is to predict the next token).

An arbitrary task can be converted into a natural language task specification, often called a \textit{prompt}. Prompting a task in this way makes its format similar to the language modeling objective used to pre-train large language models. In the zero-shot setting, this prompt contains just the task with instructions, whereas in the few-shot setting, the prompt contains both the task and several example demonstrations. When a language model is tasked to generate text to complete this prompt, it can perform the task in the process. The broader paradigm of reframing all tasks as text generation is known as \textit{prompt-based learning}. In the few-shot setting, the learning that occurs from examples provided in a given prompt (the context) is known as \textit{in-context learning} \citep{promptingsurvey}. In the zero-shot setting, models perform \textit{instruction following} \citep{instructgpt}, with their performance guided through natural language instructions provided in the prompt.

Emergent prompt-based learning capabilities have mainly been demonstrated for unidirectional language models. Bidirectional language models have stronger learned representations \citep{bert,xlm,t5}; however, they have not been able to broadly demonstrate the same few-shot in-context learning capabilities or zero-shot instruction following capabilities due to the incompatibility bidirectional denoising pre-training objectives have with the prompting paradigm. Instead, they typically require fine-tuning. Bidirectional models are not able to generate long, fluent completions to prompts since they are usually only trained to output single tokens or short spans of text to in-fill masked tokens during pre-training. We discuss this more in-depth in Section \ref{sec:directionality}.

Today, language model architects are faced with a difficult choice between unidirectional or bidirectional models. The authors of GPT-3 lay out this design dilemma in \citet{gpt3}:
\begin{displayquote}
 \small{``GPT-3 has several structural and algorithmic limitations ... as a result our experiments do not include any bidirectional
architectures or other training objectives such as denoising ... our design decision comes at the cost of potentially worse performance on tasks
which empirically benefit from bidirectionality ... making a bidirectional model at the scale of GPT-3, and/or trying to make bidirectional models work with few- or zero-shot learning, is a promising direction for future research, and could help achieve the `best of both worlds'.''}
\end{displayquote}

\fewshotfigure

In this paper, we directly address this dilemma. We contribute a new technique, \textsc{Sap} (\textbf{S}equential \textbf{A}utoregressive \textbf{P}rompting), that enables bidirectional language models to take advantage of prompting and allows them to perform at the level of unidirectional models in few- or zero-shot learning without fine-tuning. \textsc{Sap} iteratively prompts bidirectional models, concatenating previous generations back into the prompt, to produce longer generations from models that were only pre-trained to output short, mask-infill spans. We acknowledge efficiency concerns in Section \ref{sec:conclusion} and we discuss the importance and impact of \textsc{Sap} and its results to the field regardless of those concerns.

Using the machine translation task as an in-depth case study, we empirically demonstrate mT5~\citep{xue-etal-2021-mt5}, a bidirectional language model, used with \textsc{Sap} outperforms its unidirectional counterparts, GPT-3 and XGLM~\citep{gpt3,xglm} in both the few-shot and zero-shot settings, while utilizing approximately 50\% fewer parameters. We then examine \textsc{Sap}'s effectiveness on other tasks such as question answering and summarization, demonstrating that bidirectional models can be prompted for tasks beyond machine translation.

Our work hints at the possibility of more efficient and performant few-shot learners through pre-trained language models that incorporate bidirectionality. We discuss this impact and outline future research directions to this end in Section \ref{sec:conclusion}. In summary, our key contributions are:
\begin{enumerate}
    \item{We introduce \textsc{Sap}, a technique that enables bidirectional language models to work with few-shot and zero-shot prompt-based learning at a level that exceeds unidirectional models. Our results demonstrate in-context learning and instruction following are emergent properties of a broader class of language models, rather than only unidirectional models, addressing a long-standing challenge in language model design and use.}
    \item{We perform an in-depth study of the effectiveness of a bidirectional language model, mT5, with \textsc{Sap} on the machine translation task. Evaluating over 14 language pairs, despite using approximately 50\% fewer parameters than GPT-3 and XGLM, we find \textsc{Sap} with mT5 has improved average few-shot and zero-shot performance over all language pairs, and especially has improved performance on individual low-resource language pairs. }
    \item{We propose a range of improvements---filtering, prompt ensembling, and English-centric bootstrapping---to the unsupervised machine translation procedure outlined by \citet{gpt3unsupervised} to better adapt the bootstrapping process for unsupervised low-resource machine translation.}
    \item{We assess \textsc{Sap}'s performance on the tasks of question answering and summarization, and we find the technique enables few-shot in-context learning and zero-shot instruction following capabilities of bidirectional models in tasks beyond machine translation.}
\end{enumerate}

\section{Related Work}

\subsection{Unidirectional and Bidirectional Language Models}
\label{sec:directionality}
Transformer-based language models \citep{attention} can be broadly categorized into bidirectional and unidirectional models. Bidirectional models are models that use a denoising pre-training objective (such as masked language modeling), allowing them to utilize \textit{bidirectional} context when learning language representations. Unidirectional language models are models with a causal---or a left-to-right---language modeling objective (such as next token prediction), restricting them to be \textit{unidirectional} when learning representations \citep{promptingsurvey}.

The T5 family of models, such as T5 v1.1 and mT5, and BART-style models \citep{bart} are bidirectional, while GPT-style models, such as GPT-2, GPT-3, and XGLM are unidirectional.  Usually, but not always, bidirectional models are paired with an encoder-decoder architecture, while unidirectional models are paired with a decoder-only architecture \citep{bert, t5,xue-etal-2021-mt5,gpt2,gpt3,xglm,bigsciencearchobjective}. BERT-style models are an example of an exception. BERT-style models are bidirectional, but they cannot be easily utilized for prompting and text generation since they are encoder-only \citep{bertspeak}. Of the available bidirectional models, T5 models are the only models with a long enough sequence length (unlimited with their relative position embeddings) to support many in-context prompt examples and with a large enough number of parameters to be effective zero-shot and few-shot performers \citep{gpt2,gpt3,scaling}.  See Appendix \ref{sec:survey} for a survey of popular open source language models. Aside from sequence length and model size, BART is not purely trained on the span denoising objective \textsc{Sap} exploits, but is also trained on many other corruption objectives like ``Sentence Permutation.'' For this reason, we utilize the T5 models for experiments and leave the exploration of the generalization of \textsc{Sap} to other models, that could become available later, as future work.

\citet{bert} and \citet{t5} have both shown that after transfer learning, bidirectional denoising pre-training objectives such as BERT's masked language modeling and T5's random span corruption outperform causal language modeling on downstream tasks. \citet{gpt3} concedes this to be a potential source of weakness for the GPT-3 model on certain tasks where bidirectionality is important.

Despite the advantages of denoising objectives, prompting and in-context learning capabilities have not been broadly demonstrated for bidirectional language models like T5, disqualifying them when few-shot in-context learning and zero-shot instruction following is desired. \citet{lester-etal-2021-power} explains this may be because: 
\begin{displayquote}
 \small{``...a T5 model pre-trained exclusively on span corruption, such as T5.1.1, has never seen truly natural input text (free of sentinel tokens), nor has it ever been asked to predict truly natural targets''}
\end{displayquote}

In other words: when pre-trained on their denoising objectives, language models like T5 that utilize bidirectionality are only conditioned to output a single token or short spans of tokens (the in-fill of the mask) rather than full and complete sentences; this inhibits their ability to generate arbitrary-length natural responses to a variety of prompts.

Despite the stronger learned representations of bidirectional models, their shortcomings in prompt-based learning motivate \citet{gpt3} and \citet{xglm} to explicitly choose unidirectional models over bidirectional models for GPT-3 and XGLM.

\subsection{Prompting Bidirectional Language Models}
\label{sec:priortechniques}

Unlike prior approaches to incorporate prompt-based learning capabilities into bidirectional models, our technique, \textsc{Sap}, neither requires fine-tuning, weight updates, nor supervised instruction-tuning datasets. It demonstrates that bidirectional language models develop \textit{innate} few-shot learning capabilities with in-context learning and zero-shot instruction following capabilities.

\paragraph{Cloze-style prompts} \citet{cloze} and \citet{cloze2} find that bidirectional models such as RoBERTa and ALBERT \citep{roberta,albert} can be ``prompted'' with cloze-style phrases. They propose a few-shot training paradigm called \textsc{Pet} where the model's predicted mask in-fill, called a ``verbalizer,'' is used to label fine-tuning examples for the model. These verbalizers are only a single word or a few words, e.g. ``yes'', ``no'', ``amazing'', ``worse''. \citet{electrazero} follow a similar technique, but with the ELECTRA model \citep{electra}.  These works primarily demonstrate zero-shot effectiveness on classification tasks such as sentiment analysis, rather than more challenging generation tasks such as machine translation or question answering. Furthermore, they still require fine-tuning for effective few-shot learning, a major limitation that does not achieve the prompt-based in-context learning or instruction following abilities of unidirectional models such as GPT-3.

\paragraph{LM-adaptation}
\citet{lester-etal-2021-power} finds some success with prompting the T5 v1.1 models after continued pre-training on the unidirectional prefix-LM objective described in \citet{t5}. The resulting model, T5 v1.1 LM-adapted (T5+LM), is described as a late-stage adaptation to a unidirectional objective. Adaptation requires performing weight updates, and given that representations learned by the original denoising objective have been shown to be superior \citep{t5}, we hypothesize that such an adaptation could degrade the quality of the learned representations.

\paragraph{Prompt-tuning}
\citet{lester-etal-2021-power} and \citet{prefixtuning} find by fine-tuning only a portion of the parameters in an otherwise frozen pre-trained bidirectional language model, a ``soft prompt'' can be discovered through backpropagation. Soft prompts are prompts discovered in the embedding space of the model and are not grounded in natural language. As a form of parameter-efficient fine-tuning \citep{parameff}, this approach requires training the prompt embeddings and benefits from initialization from LM-adaptation, both of which require performing weight updates. The nature of soft prompts lacking grounding in natural language makes their use and flexibility limited, a stark difference from the instruction following capabilities of unidirectional models  \citep{promptingsurvey}.

\paragraph{Instruction-tuning}

Language models can be fine-tuned on a supervised dataset consisting of natural language prompts and their respective target completions \citep{flan,t0,instructgpt,metalicl}. This ``instruction-tuning'' technique allows these models to improve performance on instruction following and therefore exhibit few-shot and zero-shot capabilities through prompting. The T0 model in particular is an instruction-tuned version of the T5+LM model~\citep{lester-etal-2021-power}, augmenting it with prompting capabilities. While instruction-tuning likely bolsters the instruction following performance of a model, we hypothesize that by instruction-tuning, the T0 model is to some degree surfacing the innate prompting ability that the bidirectional model already has. We provide evidence towards this hypothesis by demonstrating that bidirectional models can be prompted without instruction-tuning.

\subsection{Unsupervised Machine Translation through Prompting}

GPT-2~\citep{gpt2}  and GPT-3~\citep{gpt3} have shown it is possible to perform few-shot machine translation and unsupervised zero-shot machine translation with large language models using prompting and in-context learning. The XGLM model~\citep{xglm} trains a similar architecture to GPT-3 on a diverse multilingual corpus, resulting in improvements on few-shot, low-resource machine translation. \citet{gpt3unsupervised} introduce bootstrapping and self-amplification techniques to further improve unsupervised zero-shot performance on machine translation.

\firstwordablationtable
\section{Few-shot Machine Translation}
\label{sec:few-shot}

To motivate our method for enabling few-shot in-context learning in bidirectional language models, we first focus on applying $\text{mT5}_\text{3.7B}$ (mT5-XL) \citep{xue-etal-2021-mt5} to the machine translation task as an in-depth case study since this task benefits greatly from bidirectionality \citep{xlm,xglm}. We largely follow the procedure of \citet{xglm}, except with mT5 and \textsc{Sap}. mT5 is a massively multilingual bidirectional model trained on random span corruption, a variant of masked language modeling.  We demonstrate that with \textsc{Sap}, mT5 can perform few-shot machine translation using prompting and in-context examples with no fine-tuning. We first formulate a prompt format that utilizes its random span masking scheme to complete the translation task, such as:
\\[.5em]\centerline{\fbox{\begin{minipage}{15.5em}
\small{
Translate Spanish to English.\\
Spanish: El clima es soleado.\textcolor{gray}{</s>}\\
English: The weather is sunny.\textcolor{gray}{</s>}\\
Spanish: Mi perro es un cachorro.\textcolor{gray}{</s>}\\
English: My dog is a puppy.\textcolor{gray}{</s>}\\
Spanish: Los árboles son importantes.\textcolor{gray}{</s>}\\
English: \textcolor{red}{<X>}
}
\end{minipage}}}

\subsection{Sequential Autoregressive Prompting (\textsc{Sap}) Technique}

By requiring mT5 to in-fill \textcolor{red}{<X>}\footnote{We use the first sentinel token from the mT5 vocabulary as our mask token.}, we are effectively asking it to translate the Spanish sentence. However, due to the limitations of the denoising pre-training objective on prompting (described in Section \ref{sec:directionality}), we observe mT5 often outputs a partial translation of the beginning of the source sentence, rather than the full translation. To overcome this, we prompt mT5 $T$ times until the model generates a stop token \textcolor{gray}{</s>}\footnote{We repurpose the 100th sentinel token from the mT5 vocabulary as our stop token.}, resulting in a longer translation. At each time step of iteration, we keep the first word generated (using the space character as delimiter) and concatenate it into the last line of the prompt to use in the next time step. This iterative prompting enables us to extract longer generations. Formally, we denote the generation at each time step $t$ as $G_t$. We denote the first word generated at each time step as $F_t$, where $F_t = \texttt{SPLIT}(G_t, \texttt{" "})\texttt{[0]}$. We update the prompt at each time step $P_t$ to include the cumulative generation from all previous time steps concatenated in the last line of the prompt. The prompt used at each time step $P_t$ is as follows:
\\[.5em]\centerline{\fbox{\begin{minipage}{15.5em}
\small{
Translate Spanish to English.\\
Spanish: El clima es soleado.\textcolor{gray}{</s>}\\
English: The weather is sunny.\textcolor{gray}{</s>}\\
Spanish: Mi perro es un cachorro.\textcolor{gray}{</s>}\\
English: My dog is a puppy.\textcolor{gray}{</s>}\\
Spanish: Los árboles son importantes.\textcolor{gray}{</s>}\\
English: \texttt{CONCAT}($F_0$, \ldots,  $F_{t-1})$
\textcolor{red}{<X>}
}
\end{minipage}}}\\[.5em]
In Table \ref{table:firstword-ablation}, we also consider sequential prompting---concatenating the entire generation $G_t$ instead of just the first word of the generation $F_t$---but find that it produces significantly inferior results as low-quality tokens are generated after the first word. By conditioning the model to generate the next word in the translation based on previous words generated, this technique resembles autoregression. mT5 is already autoregressive, but it is autoregressive only at the decoder level. Adding previously generated words back into the prompt allows them to pass through the encoder layers as well. For this reason, we call this technique \textsc{Sap} (\textbf{S}equential \textbf{A}utoregressive \textbf{P}rompting). To provide a signal to stop generation, we add our stop token at the end of each example in the prompt.  We stop prompting after the model generates a stop token.\footnote{We also implement a basic post-processing step to strip any generated text after a repeated sequence of three or more tokens following settings available in common decoding implementations \citep{transformers}.} The overall process is graphically depicted, with stop tokens omitted, in Figure \ref{fig:fewshot}.

\subsection{Results}
\label{sec:fewshot-results}
Following \citet{xglm}, we evaluate our technique on 14 languages from the FLORES-101 dataset \citep{flores101} that span high-resource and low-resource languages\footnote{HR: English (en), German (de), French (fr), Catalan (ca), Finish (fi), Russian (ru), Bulgarian (bg), Chinese (zh), Korean (ko), Arabic (ar), Swahili (sw); LR: Hindi (hi), Malayalam (my), Tamil (ta)}. We evaluate SentencePiece BLEU (spBLEU) \citep{flores101} in every direction, leading to an evaluation over 182 language pairs in total. Abbreviated results can be found in Table \ref{table:fewshot-flores-results-abbrev}, and the matrix of full results can be found in Appendix \ref{sec:flores-results}. Examples generations can be found in Appendix \ref{sec:examplegenerations}.

On an average spBLEU score over all 182 pairs, our model matches the performance of the unidirectional XGLM and GPT-3 models---with approximately 50\% fewer parameters and 16x fewer examples. Notably, our technique has significant improved performance on language pairs with at least one low-resource language, while trailing only slightly on high-resource pairs.

\bootstrapfigure

\section{Unsupervised Zero-shot Machine Translation}
\label{sec:zero-shot}

To extend our in-depth case study on the machine translation task, we now perform fully unsupervised zero-shot machine translation with \textsc{Sap} and mT5 following the procedure of \citet{gpt3unsupervised}, which uses a self-amplification technique to boost performance. A comparison of zero-shot performance without self-amplification can be found in Appendix \ref{sec:selfamp-ablation}. We ultimately will replace the examples in the few-shot prompt with synthetic parallel examples. These synthetic parallel examples are bootstrapped in a completely unsupervised fashion using a zero-shot translation prompt with no examples. The zero-shot prompt format looks like:
\\[.5em]\centerline{\fbox{\begin{minipage}{15.5em}
\small{
Translate Spanish to English.\\
Spanish: Los árboles son importantes.\textcolor{gray}{</s>}\\
English: \textcolor{red}{<X>}
}
\end{minipage}}}\\[.5em]
We adapt the bootstrap process of \citet{gpt3unsupervised} to retrieve these synthetic parallel examples. The process, as depicted in Figure \ref{fig:bootstrap}, consists of three steps:

\begin{itemize}[leftmargin=*]
\item[] \textbf{Step 1 (sampling)}: Generate synthetic parallel examples using a zero-shot translation prompt (with no examples) to translate sentences from a monolingual source language corpus.

\item[] \textbf{Step 2 (filtering)}: Filter out low-quality synthetic examples to keep only high-quality synthetic examples using an unsupervised scoring technique (discussed in Section \ref{sec:mt5score}).

\item[] \textbf{Step 3 (self-amplification)}: Translate any source language sentence desired using these synthetic parallel examples in the few-shot prompt.

\end{itemize}

We iteratively run multiple rounds of this bootstrap by repeating step 2 and step 3 to form a better few-shot prompt. The few-shot prompt after self-amplification is used to translate more source language sentences. These are then filtered using the scoring technique used in step 2 and so on. In our experiments, we run four bootstrapping rounds and sample 100 source language sentences from the training dataset in each round. Note that none of the target language parallel sentences from the training dataset are used in this zero-shot setting; following \citet{gpt3unsupervised}, only the source language sentences are used.

\subsection{Filtering Down to High-quality Translations}
\label{sec:mt5score}

The filtering step of the bootstrap requires an unsupervised scoring method for assessing the quality of translations. We first use \texttt{{langdetect}}\footnote{\url{https://pypi.org/project/langdetect/}}, a language identifier,  as a simple rule-based filter to ensure the generated text is in the desired target language. We then score the remaining generated translations against their corresponding original sentence in the source language. For this unsupervised multilingual similarity metric, we utilize the BERTScore~\citep{bertscore} algorithm with $\text{mT5}_{\text{300M}}$ (mT5-small)\footnote{The BERTScore Python library~\citep{bertscore} directly supports using mT5 instead of BERT.}, dubbing it ``mT5Score''. We ablate the use of mT5Score as a filter in Appendix \ref{sec:mt5score-ablation}.

We take the top two synthetic parallel examples with the highest mT5Score in the filtering step and use those as synthetic few-shot examples in the prompt in the self-amplification step.

\subsection{Translating with an Ensemble of Prompts}

Because the two examples used in the prompt can greatly affect the quality of the generated translations, some prompts containing low-quality synthetic examples may cause poor translations for certain sentences. To combat this and reduce variation in performance,  we keep the top $N$ synthetic examples instead of two synthetic examples. We use these to form $\frac{N}{2}$ different few-shot prompts with two synthetic parallel examples each. Each sentence in the test set is then translated with these $\frac{N}{2}$ different prompts to produce $\frac{N}{2}$ translations. The best translation of the $\frac{N}{2}$ translations is chosen in a fully unsupervised manner with mT5Score, as done in the filtering step of the bootstrap.

We find this ensembling technique helps make unsupervised zero-shot performance competitive with few-shot performance. Experiments varying the number of prompts in the ensemble can be found in Appendix \ref{sec:promptensemble-ablation}. Unless otherwise stated, we use a 4 prompt ensemble in this paper: $\frac{N}{2} = 4$.

In sum, we sample and zero-shot translate 100 sentences from a monolingual corpus, keep the top eight synthetic parallel examples scored by mT5Score, and use them to form four few-shot prompts, each of which has two synthetic examples. 

\fewshotfloresresultstableabbrev
\subsection{English-centric Bootstrapping}
\label{sec:englishcentric}

While \citet{gpt3unsupervised} only performed a bootstrap on English-French and French-English pairs, we perform bootstrapping on some language pairs which may contain at least one low-resource language or non-English language.

It has been found that multilingual language models perform best in English,  due to  imbalance of languages in the pre-training corpus where English has the highest amount of data \citep{xglm}. Therefore, when running the bootstrap on various language pairs, we modify the bootstrap to favor generating English, or pivot through English when neither the source nor target language is English. Ablation experiments can be found in Appendix \ref{sec:englishcentric-ablation}. We outline examples of our modified English-centric bootstrapping process for various language pairs in Appendix \ref{sec:englishcentric-examples}.

\subsection{Results}
We report results with the same method used for our few-shot evaluation. Abbreviated results can be found in Table \ref{table:fewshot-flores-results-abbrev} and the matrix of full results can be found in Appendix \ref{sec:flores-results}.

In this unsupervised setting, we find our zero-shot results exceed our 2-shot results; furthermore, they significantly exceed the performance of the XGLM and GPT-3 results reported in \citet{xglm} on an average spBLEU score over all 182 pairs (+1.0 spBLEU). Again, we note strong performance on language pairs that contain one or more low-resource languages.

Intuitively, we can explain the zero-shot performance surpassing the few-shot performance through our use of prompt ensembling in the zero-shot setting. As prompt ensembling utilizes four prompts with two synthetic parallel examples each, it essentially uses eight synthetic examples, instead of just two real examples in the few-shot setting. Our synthetic examples are nearly as high-quality as real examples (similar to the findings of \citet{gpt3unsupervised}) as demonstrated by Appendix \ref{sec:promptensemble-ablation}. Prompt ensembling not only reduces performance variation if low-quality synthetic examples are selected during the bootstrap, but it also boosts performance beyond the few-shot setting as demonstrated by Table \ref{table:firstword-ablation} and the Appendix \ref{sec:promptensemble-ablation} experiments (Russian-English 26.9 $\rightarrow$ 27.9 spBLEU).

In Appendix \ref{sec:zeroshot-results}, we also evaluate on WMT14~\citep{wmt14} to compare with the results reported in \citet{gpt3unsupervised} using $\text{GPT-3}_\text{175B}$.

\section {Other Language Generation Tasks}
\label{sec:other-tasks}

We next demonstrate that bidirectional models have a generalized ability, beyond machine translation, to be prompted for arbitrary tasks. We evaluate their performance on question answering and summarization language generation tasks. Example generations can be found in Appendix \ref{sec:examplegenerations}.

\subsection{Question Answering}

We compare the zero-shot question answering performance of mT5 against XGLM on the XQuAD dataset \citep{xquad}, a multilingual question answering dataset, in Table \ref{table:xquad}. We find mT5 with \textsc{Sap} outperforms XGLM significantly (+1.7 EM, +12.3 F1). 

In Table \ref{table:squad}, we also compare against T5+LM~\citep{lester-etal-2021-power}. As T5+LM is English-only, we compare using the English-only SQuAD v1.1 dataset \citep{squad}. We still utilize the multilingual mT5 with \textsc{Sap} due to observations that the English-only T5 v1.1 model does not perform as well as mT5 in prompt-based learning\footnote{We discuss this observation in more detail in Appendix \ref{sec:t5v11observation}.}. \textsc{Sap} achieves +6.7 EM and +5.6 F1 over T5+LM.

\textsc{Sap}, as an iterative technique, is useful for producing long generations from a bidirectional model for tasks such as machine translation. We find, however, it still has utility on tasks like question answering where answer generations are shorter spans of text. We ablate utilizing \textsc{Sap} with mT5 against the simple approach of prompting mT5 once and using the mask in-fill generated on SQuAD v1.1. In the few-shot (16-shot) setting, we find that utilizing \textsc{Sap} still markedly improves performance (+12.5 EM, +5.5 F1) even on short-form generation tasks like question answering.

\xquadtable
\squadandsummarizationtable

\subsection{Summarization}

We next perform summarization on the CNN/Daily Mail dataset \citep{cnndailymail,cnndailymail2,cnndailymail3} as another long-form text generation task. We compare mT5 with T5+LM and ablate the usage of \textsc{Sap} once again in Table \ref{table:summarization}. In the few-shot setting, we find a significant lead against T5+LM (+7.1 ROUGE-L). Of that +7.1 ROUGE-L boost, the ablation of our usage of \textsc{Sap} finds the technique itself is responsible for a large component of the boost (+5.3).

\section {Conclusion and Future Directions}
\label{sec:conclusion}

We demonstrate \textsc{Sap} with the bidirectional mT5 model enables few-shot and zero-shot machine translation and zero-shot multilingual question answering, outperforming unidirectional models despite using far fewer parameters and examples. Our results suggest that the bidirectional representations learned by models such as mT5 contribute to this improved performance. Still, we concede that our results do not conclusively prove bidirectionality explains the difference in performance. Beyond bidirectionality and pre-training objectives, mT5, XGLM, and GPT-3 further differ in architecture, pre-training corpus, and hyperparameters. A complete ablation experiment would be computationally expensive, and we leave this as future work. The main limitation of \textsc{Sap} lies in its computational efficiency, discussed further in Appendix \ref{sec:limitations} along with potential mitigations.

Importantly, these results demonstrate bidirectional models possess few-shot in-context learning and zero-shot instruction following capabilities innately, without the post-hoc modifications required by prior work. Our results finally contribute strong evidence towards the strength and efficiency of bidirectional pre-training objectives and motivate further research into bidirectional architectures, pre-training objectives, and language models designed and optimized for prompting and few-shot learning. We hypothesize these future bidirectional training schemes could yield an approach that overcomes the efficiency limitations of \textsc{Sap}, while maintaining the performance and parameter size reduction benefits. Concurrent recent work that compares or mixes unidirectional and bidirectional pre-training objectives \citep{bigsciencearchobjective,unifying,alexatm} already provide some early evidence towards this hypothesis.

\subsubsection*{Acknowledgments}
We thank Daphne Ippolito for reviewing versions of this draft and Victor Sanh for answering queries related to earlier directions of this work. This research is based upon work supported in part by the DARPA KAIROS Program (contract FA8750-19-2-1004), the DARPA LwLL Program (contract FA8750-19-2-0201), the IARPA BETTER Program (contract 2019-19051600004), the IARPA HIATUS Program (contract 2022-22072200005), and the NSF (Award 1928631). Approved for Public Release, Distribution Unlimited. The views and conclusions contained herein are those of the authors and should not be interpreted as necessarily representing the official policies, either expressed or implied, of ODNI, DARPA, IARPA, NSF, or the U.S. Government.

\bibliography{anthology,iclr2023_conference}
\bibliographystyle{iclr2023_conference}

\appendix
\nopagebreak
\section{FLORES-101 Few-shot and Unsupervised Zero-shot Machine Translation}
\label{sec:flores-results}

\fewshotfloresresultstable
\raggedbottom

\section{WMT14 Unsupervised Zero-shot Machine Translation}
\label{sec:zeroshot-results}

\selfamplificationtable

\section{Filtering and Selection Ablation}
\label{sec:mt5score-ablation}
\mtscoreablationtable

\section{Prompt Ensemble Size}
\label{sec:promptensemble-ablation}

\promptensembleablationtable

\section{English-centric Bootstrap Ablation}
\label{sec:englishcentric-ablation}

\englishcentricablationtable

\section{English-Centric Bootstrap Examples}
\label{sec:englishcentric-examples}
We outline examples of our modified English-centric bootstrapping process for various language pairs below:
\begin{itemize}[leftmargin=*]
    \item \textbf{Example 1} (Russian-English): No change.
    \item \textbf{Example 2} (English-Russian): In step 1, generate Russian-English synthetic examples using a Russian monolingual corpus. Then, reverse the examples to obtain English-Russian synthetic examples. 
    \item \textbf{Example 3} (Russian-Chinese): In step 1, for the first three rounds of the bootstrap, generate Russian-English synthetic examples and Chinese-English synthetic examples using Russian and Chinese monolingual corpora. On the fourth and final round, use an English monolingual corpus along with the reversed previous synthetic examples to produce English-Russian and English-Chinese synthetic examples. Since the same English sentences are used to produce both sets, we can align these to form synthetic Russian-Chinese examples. In step 2, we filter examples using the harmonic mean of the two mT5Scores.
\end{itemize}

\section{Zero-shot Performance Without Self-amplification}
\label{sec:selfamp-ablation}

\selfampablationtable

\section{Prompting T5 v1.1 with \textsc{Sap}}
\label{sec:t5v11observation}

Careful readers may ask why we use a multilingual model, mT5, to obtain results for the English-only tasks of QA (SQuAD) and summarization (CNN/DailyMail). While a suitable English-only version of T5 could in theory improve performance, we found issues with T5 v1.1's performance. We choose to run \textsc{Sap} with mT5 due to the observation that T5 v1.1 cannot be prompted as easily as mT5, and thus underperforms.

The inputs seen by T5 v1.1 and mT5 during pre-training are of sequence length 512 tokens where multiple spans in the sequence are dropped \citep{t5}. Therefore, the prompt template we describe in Section \ref{sec:few-shot}, would be out-of-distribution from the pre-training inputs since it may have a sequence length shorter or longer than 512 tokens and only contains a single mask instead of multiple masks.

We find that the mT5 model has generalized to sequences shorter and longer than 512 tokens and to sequences that only contain a single mask, while the T5 v1.1 model has not. It is still possible to prompt the T5 v1.1 model with \textsc{Sap}, but requires formulating a prompt constrained to the same in-distribution length of 512.

Due to this complication, we forgo prompting T5 v1.1 in this paper. Since mT5 and T5 v1.1 were trained identically (the same model architecture and hyperparameters), apart from mT5 being pre-trained on the multilingual mC4 dataset instead of the primarily English C4 dataset, we further hypothesize that this difference between T5 v1.1 and mT5 may be an artifact of which checkpoint is selected after pre-training or the length of pre-training \citep{xue-etal-2021-mt5,t5}.

\raggedbottom
\pagebreak

\section{Limitations}
\label{sec:limitations}

\textsc{Sap} requires $T$ total forward passes to produce a generation instead of a single forward pass, where $T$ equals the number of words in the generation before reaching a stop token. For example, to produce a translation that has 14 words, \textsc{Sap} requires 14 inferences of the bidirectional model. For tasks with shorter generations with only a few words, such as multilingual question answering, \textsc{Sap} is more practical, especially since it uses fewer parameters. Depending on the size of inference data, \textsc{Sap} as an inference-only prompting technique may be faster and easier to implement than methods that require fine-tuning. While these inferences must be performed sequentially due to the autoregressive nature of the technique, utilizing batching over a test set can still ensure maximum GPU utilization, which is how our experiments were performed. For longer generation tasks, we believe \textsc{Sap} is prohibitively computationally expensive and it likely would not be suitable for use by practitioners directly despite some evidence of improvements in performance.  Nevertheless, \textsc{Sap} uncovers an important result: prompt-based learning is an emergent property of bidirectional models. We hypothesize that further research into pre-training objectives and language model design following \citet{bigsciencearchobjective}, \citet{unifying}, and \citet{alexatm} could yield a bidirectional pre-training objective better optimized for few-shot prompting, lifting the requirement to perform multiple forward passes sequentially to generate longer completions. 

\raggedbottom
\pagebreak

\section{Survey of Open Source Language Models}
\label{sec:survey}

\begin{table}[H]
\fontsize{8}{12}\selectfont
\begin{tabularx}{\textwidth}{p{2.4cm}XXp{1.8cm}p{3.3cm}}
\toprule[\heavyrulewidth]
\textbf{Model} & \textbf{Architecture} & \textbf{Large \newline (>1B params?)} & \textbf{Max Sequence\newline Length during Pre-training} & \textbf{Pre-training\newline Objective} \\ \hline
\multicolumn{5}{l}{\textit{Unidirectional Pre-training Objectives}} \\ \hline
{\textbf{GPT-family models}\newline\newline(GPT-2, GPT-3)\newline{\tiny\citep{gpt2,gpt3}}} & Decoder-only & \cmark & 1024-2048 & {Next Token Prediction} \\ \hline
{\textbf{EleutherAI-family models}\newline\newline(GPT-Neo, \newline GPT-J, GPT-NeoX)\newline{\tiny\citep{gpt-neo,gptj,gpt-neo-x}}} & Decoder-only & \cmark & 2048 & {Next Token Prediction} \\ \hline
{\textbf{XGLM}\newline\newline{\tiny\citep{xglm}}} & Decoder-only & \cmark & 2048 & {Next Token Prediction} \\ \hline
{\textbf{OPT}\newline\newline{\tiny\citep{opt}}} & Decoder-only & \cmark & 2048 & {Next Token Prediction} \\ \hline
{\textbf{BLOOM}\newline\newline{\tiny\citep{bloom}}} & Decoder-only & \cmark & 2048 & {Next Token Prediction} \\ \hline
\multicolumn{5}{l}{\textit{Bidirectional Pre-training Objectives}} \\ \hline
{\textbf{BERT-style models}\newline\newline(BERT, RoBERTa, ALBERT, etc.)\newline{\tiny\citep{bert,roberta,albert}}} & Encoder-only & \xmark & 512 & {Masked Language Modeling; \newline Next Sentence Prediction} \\ \hline
{\textbf{BART-style models}\newline\newline(BART, 
\newline mBART, etc.)\newline{\tiny\citep{bart,mbart}}} & Encoder-Decoder & \xmark & 512/512 & {Token Masking; \newline Token Deletion; \newline  Sentence Permutation; \newline  Document Rotation;  \newline  Text Infilling} \\ \hline
{\textbf{T5-style models}\newline\newline(T5,  mT5, etc.)\newline{\tiny\citep{t5,xue-etal-2021-mt5}}} & Encoder-Decoder & \cmark & 1024/512 with \newline Relative \newline Position \newline Embeddings & {Random Span Corruption} \\
\bottomrule[\heavyrulewidth]
\end{tabularx}
\end{table}

\raggedbottom
\pagebreak

\section{Selected Example Generations}
\label{sec:examplegenerations}

\selectedgenerations

\pagebreak
\section{Resources}

\noindent We provide links and citations to resources used in this paper which provide license information, documentation, and their intended use. Our usage follows the intended usage of all resources.

~\\ ~\\ \noindent We utilize the following models:
\begin{itemize}
    \item mT5 \citep{xue-etal-2021-mt5}: \\ {\small\url{https://github.com/google-research/multilingual-t5/}}
    \item T5 v1.1 \citep{t5,lester-etal-2021-power}: \\ {\small\url{https://github.com/google-research/text-to-text-transfer-transformer/}}
    \item T5+LM \citep{t5,lester-etal-2021-power}: \\ {\small\url{https://github.com/google-research/text-to-text-transfer-transformer/}}

\end{itemize}

~\\ \noindent We utilize the following datasets:
\begin{itemize}
    \item FLORES-101 \citep{flores101}: \\ {\small\url{https://ai.facebook.com/research/publications/the-flores-101-evaluation-benchmark-for-low-resource-and-multilingual-machine-translation}}
    \item WMT14 \citep{wmt14}: \\
    {\small\url{https://www.statmt.org/wmt14/translation-task.html}}
    \item XQuAD \citep{xquad}: \\ {\small\url{https://github.com/deepmind/xquad}}
    \item SQuAD v1.1 \citep{squad}: \\ {\small\url{https://rajpurkar.github.io/SQuAD-explorer/}}
    \item CNN / Daily Mail v3.0.0 \citep{cnndailymail,cnndailymail2,cnndailymail3}: \\ {\small\url{https://huggingface.co/datasets/ccdv/cnn_dailymail}}
\end{itemize}

~\\ \noindent We utilize the following software:
\begin{itemize}
    \item Transformers \citep{transformers}: \\ {\small\url{https://github.com/huggingface/transformers}}
    \item Datasets \citep{lhoest-etal-2021-datasets}: \\ {\small\url{https://github.com/huggingface/datasets}}
    \item SacreBLEU \citep{sacrebleu,flores101}: \\ {\small\url{https://github.com/ngoyal2707/sacrebleu}}
    \item ROUGE \citep{rouge}: \\ {\small\url{https://github.com/pltrdy/rouge}}
    \item BERTScore \citep{bertscore}: \\ {\small\url{https://github.com/Tiiiger/bert_score/tree/master/bert_score}}
    \item \texttt{langdetect}: \\ {\small\url{https://pypi.org/project/langdetect/}}

\end{itemize}

~\\ \noindent We estimate the total compute budget and detail computing infrastructure used to run the computational experiments found in this paper below:
\begin{itemize}
    \item 1x NVIDIA RTX A6000 / 87GB RAM / 4x CPU -- 686 hours
\end{itemize}

\end{document}

%% file: math_commands.tex
\usepackage{amsmath,amsfonts,bm}

\def\eqref#1{equation~\ref{#1}}

\def\1{\bm{1}}

\DeclareMathAlphabet{\mathsfit}{\encodingdefault}{\sfdefault}{m}{sl}
\SetMathAlphabet{\mathsfit}{bold}{\encodingdefault}{\sfdefault}{bx}{n}



%% file: resources/figures.tex
\newcommand{\fewshotfigure}{
    \begin{figure}[t]
        \centering
        \includegraphics[width=400px]{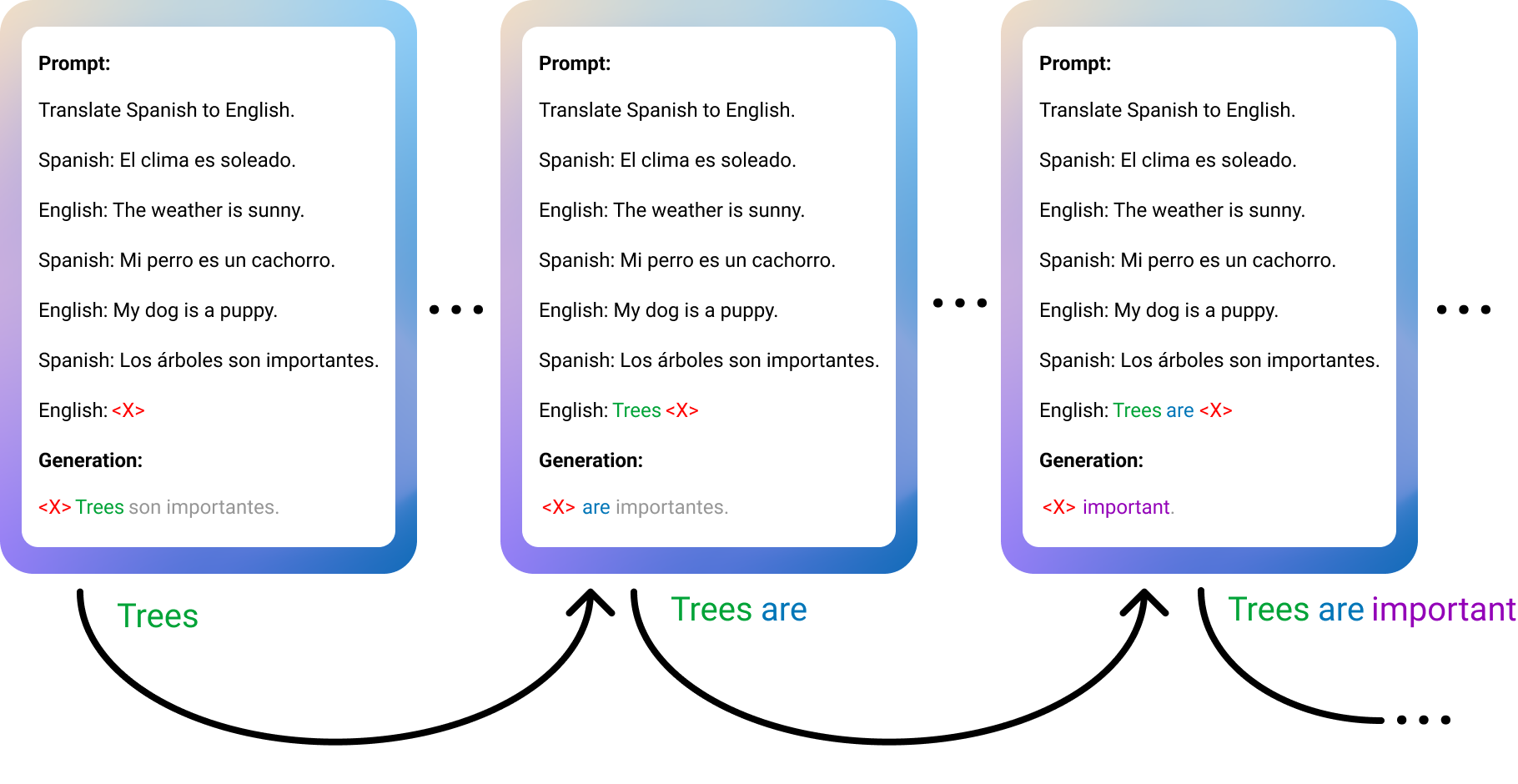}
        \caption{A visualization of our \textsc{Sap} technique extracting high-quality translations from mT5. In the zero-shot setting, the examples used in the prompt are synthetic examples retrieved in a fully unsupervised manner.}
        \label{fig:fewshot}
    \end{figure}
}

\newcommand{\bootstrapfigure}{
    \begin{figure*}
        \centering
        \includegraphics[width=400px]{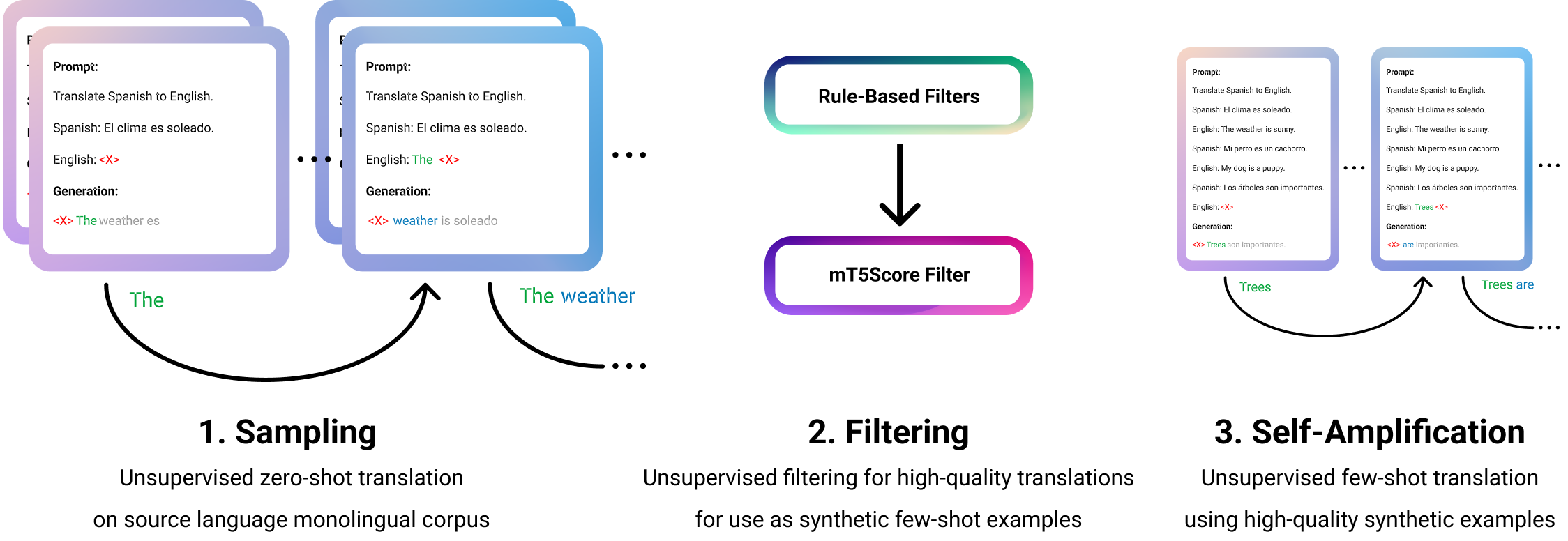}
        \caption{A visualization of the bootstrapping process described in Section \ref{sec:zero-shot}. }
        \label{fig:bootstrap}
    \end{figure*}
}

%% file: resources/few-shot_results.tex
\newcommand{\hlinep}{
    \hline & \\[-1.5ex]
}

\newcommand{\fewshotfloresresultstablerow}[7]{
    \multirow{5}{*}{#1}  & Supervised &                                \xintFor ##8 in #2 \do {  & ##8 }    \\ 
                         & $\text{GPT-3}_{\text{6.7B}}$ & (32\text{-shot})   \xintFor ##8 in #3 \do {  & ##8 }    \\
                         & $\text{XGLM}_{\text{7.5B}}$ & (32\text{-shot})    \xintFor ##8 in #4 \do {  & ##8 }    \\
                         & $\text{mT5}_{\text{3.7B}}\ + \  \text{\textsc{Sap}}$ & (2\text{-shot})      \xintFor ##8 in #5 \do {  & ##8 }    \\
                         & $\text{mT5}_{\text{3.7B}}\ + \  \text{\textsc{Sap}}$ &  (zero\text{-shot})  \xintFor ##8 in #6 \do {  & ##8 }  \\
                         #7
}

\newcommand{\fewshotfloresresultstable}{
    \begin{table}[H]
    \centering
    \fontsize{6}{6}\selectfont
    \setlength{\tabcolsep}{3pt}
    \begin{tabular}{llrcccccccccccccc|c}
    \toprule[\heavyrulewidth]
                         & &                                 & en & de & fr & ca & fi & ru & bg & zh & ko & ar & sw & hi & my & ta & avg \\ \hlinep
    \fewshotfloresresultstablerow{en}
        {{–,32.6,42.0,31.2,24.2,27.1,37.4,19.3,18.5,17.9,26.9,28.1,3.5,3.4,24.0}}
        {{–,25.9,\textbf{36.1},23.8,10.2,11.2,5.9,12.5,1.2,1.1,0.5,0.3,0.1,0.0,9.9}}
        {{–,\textbf{27.6},36.0,\underline{\textbf{34.0}},\textbf{23.3},\textbf{24.2},\textbf{33.1},\textbf{15.6},\textbf{12.0},11.5,18.0,\textbf{19.9},\underline{11.0},\underline{8.5},\textbf{21.1}}}
        {{–,23.2,34.2,26.2,15.8,20.1,27.9,9.5,10.4,11.4,17.3,14.0,\underline{11.0},\underline{11.2},17.9}}
        {{–,26.0,33.2,28.4,15.7,21.2,27.1,11.3,10.5,\textbf{12.7},\textbf{19.1},16.1,\underline{\textbf{13.2}},\underline{\textbf{13.1}},19.0}}{\hlinep}

    \fewshotfloresresultstablerow{de}
        {{35.8,–,35.5,25.8,22.6,24.6,31.5,17.2,16.6,14.8,21.0,23.4,2.3,2.3,21.0}}
        {{\underline{\textbf{40.4}},–,26.2,17.2,8.1,9.3,4.8,9.0,1.0,0.9,0.5,0.3,0.1,0.1,9.1}}
        {{\underline{38.8},–,\textbf{27.9},19.1,\textbf{20.5},\textbf{19.7},\textbf{25.8},\textbf{12.3},3.4,6.6,11.7,\textbf{14.3},\underline{\textbf{9.9}},\underline{4.8},\textbf{16.5}}}
        {{33.0,–,24.4,17.8,14.1,15.7,20.2,8.2,\textbf{9.1},7.7,11.0,10.0,\underline{9.8},\underline{9.6},14.7}}
        {{\underline{35.9},–,25.9,\textbf{22.5},14.3,17.4,21.0,8.2,8.4,\textbf{8.7},\textbf{13.4},10.4,\underline{9.0},\underline{\textbf{10.8}},15.8}}{\hlinep}

    \fewshotfloresresultstablerow{fr}
        {{37.2,28.5,–,28.7,21.9,24.5,32.2,17.6,16.7,15.4,17.2,22.9,2.1,0.8,20.4}}
        {{\underline{\textbf{42.8}},20.9,–,23.7,8.0,9.7,4.6,9.1,1.0,1.0,0.4,0.3,0.1,0.0,9.4}}
        {{\underline{40.4},20.4,–,\underline{\textbf{32.1}},\textbf{19.4},\textbf{19.8},\textbf{26.3},\textbf{10.6},2.4,5.9,14.5,\textbf{13.7},\underline{9.7},\underline{6.6},\textbf{17.1}}}
        {{\underline{38.0},19.2,–,26.7,13.7,18.3,23.5,8.6,\textbf{9.2},9.9,15.0,12.1,\underline{\textbf{10.8}},\underline{9.7},16.5}}
        {{\underline{38.1},\textbf{21.1},–,\underline{30.1},12.9,18.1,22.3,8.7,\textbf{9.2},\textbf{11.1},\textbf{15.7},11.0,\underline{9.6},\underline{\textbf{11.1}},16.8}}{\hlinep}

    \fewshotfloresresultstablerow{ca}
        {{33.4,24.8,35.1,–,19.0,21.1,28.6,15.1,13.9,13.4,18.7,20.5,2.1,2.6,19.1}}
        {{\underline{40.2},18.6,31.4,–,7.0,9.3,4.3,8.0,0.9,0.9,0.3,0.4,0.1,0.1,9.3}}
        {{\underline{\textbf{41.1}},18.9,\textbf{33.8},–,11.3,3.3,\textbf{23.9},\textbf{10.8},1.3,0.8,13.8,6.1,\underline{7.9},\underline{3.1},13.6}}
        {{\underline{33.4},14.9,29.5,–,10.7,14.0,15.6,6.5,7.0,5.6,12.4,7.3,\underline{\textbf{8.7}},\underline{6.7},13.3}}
        {{\underline{37.1},\textbf{19.3},32.4,–,\textbf{12.4},\textbf{16.7},19.1,7.9,\textbf{7.4},\textbf{8.5},\textbf{14.5},\textbf{9.4},\underline{8.3},\underline{\textbf{9.8}},\textbf{15.6}}}{\hlinep}

    \fewshotfloresresultstablerow{fi}
        {{27.2,23.0,29.3,21.6,–,20.6,26.4,16.0,14.8,12.4,14.2,19.8,1.7,0.9,17.5}}
        {{25.3,13.5,17.1,10.0,–,6.4,2.8,5.7,0.7,0.7,0.3,0.3,0.1,0.0,6.4}}
        {{\underline{\textbf{29.2}},\textbf{17.4},\textbf{22.2},\textbf{17.0},–,\textbf{16.5},\textbf{17.5},\textbf{12.4},\textbf{7.5},\textbf{7.6},8.0,\textbf{10.1},\underline{6.2},\underline{2.0},\textbf{13.4}}}
        {{24.1,16.1,19.8,14.9,–,14.2,17.0,7.0,5.8,7.1,8.3,5.6,\underline{\textbf{8.5}},\underline{3.9},11.7}}
        {{23.2,16.1,20.5,16.3,–,14.5,16.3,8.0,5.9,6.3,\textbf{10.0},7.5,\underline{5.9},\underline{\textbf{8.2}},12.2}}{\hlinep}

    \fewshotfloresresultstablerow{ru}
        {{27.5,23.5,30.1,22.0,19.4,–,31.0,16.5,15.3,13.5,18.1,20.9,2.2,2.3,18.6}}
        {{\underline{28.1},14.8,20.4,13.1,5.4,–,7.4,1.2,0.2,0.2,0.1,0.2,0.1,0.1,7.0}}
        {{\underline{\textbf{30.4}},\textbf{17.9},\textbf{24.0},14.6,8.0,–,\textbf{26.3},\textbf{11.6},5.5,7.4,7.1,\textbf{9.1},\underline{7.3},\underline{3.1},13.2}}
        {{26.9,16.6,22.4,14.5,11.2,–,25.2,6.1,8.0,6.4,11.3,\textbf{9.1},\underline{\textbf{9.8}},\underline{8.4},13.5}}
        {{\underline{27.9},17.1,22.5,\textbf{19.4},\textbf{13.1},–,25.4,8.3,\textbf{8.7},\textbf{9.1},\textbf{12.0},9.0,\underline{9.0},\underline{\textbf{10.3}},\textbf{14.8}}}{\hlinep}

    \fewshotfloresresultstablerow{bg}
        {{33.0,26.1,33.7,24.9,20.8,26.5,–,17.5,16.4,14.5,20.9,23.1,2.3,2.4,20.2}}
        {{21.6,11.4,16.0,9.7,4.3,6.5,–,1.2,0.2,0.2,0.1,0.2,0.1,0.1,5.5}}
        {{\underline{\textbf{35.5}},\textbf{19.2},\textbf{26.3},12.9,\textbf{14.2},22.9,–,\textbf{11.9},6.8,\textbf{9.2},9.4,7.5,\underline{3.2},1.0,13.9}}
        {{31.0,17.0,23.8,18.3,10.9,22.9,–,7.2,\textbf{8.3},8.1,11.7,7.4,\underline{\textbf{9.5}},\underline{6.6},14.1}}
        {{32.5,17.3,24.5,\textbf{21.7},10.6,\textbf{23.2},–,8.7,7.5,9.0,\textbf{13.0},\textbf{8.6},\underline{7.9},\underline{\textbf{10.1}},\textbf{15.0}}}{\hlinep}

    \fewshotfloresresultstablerow{zh}
        {{20.9,17.6,24.3,17.4,16.0,17.2,22.1,–,15.9,11.6,15.5,18.5,1.9,2.5,15.5}}
        {{\underline{\textbf{21.1}},9.5,14.3,8.2,4.3,3.6,1.3,–,1.1,0.4,0.2,0.2,0.1,0.0,4.9}}
        {{20.7,8.3,8.5,10.5,4.4,4.8,\textbf{14.8},–,\textbf{9.3},4.2,5.6,\textbf{12.0},\underline{8.6},\underline{6.2},9.1}}
        {{19.0,\textbf{10.9},\textbf{14.9},11.9,8.0,10.6,11.9,–,8.9,6.0,\textbf{9.1},8.0,\underline{\textbf{10.0}},\underline{7.6},10.5}}
        {{18.5,\textbf{10.9},14.8,\textbf{12.8},\textbf{8.8},\textbf{10.7},11.8,–,9.2,\textbf{6.5},9.0,8.9,\underline{8.2},\underline{\textbf{8.9}},\textbf{10.7}}}{\hlinep}

    \fewshotfloresresultstablerow{ko}
        {{20.9,16.7,22.1,16.5,14.9,15.5,21.1,15.7,–,10.6,15.1,18.7,1.9,4.0,14.9}}
        {{8.3,4.6,6.4,4.4,2.1,1.7,0.8,2.5,–,0.2,0.1,0.1,0.1,0.1,2.4}}
        {{\textbf{19.9},\textbf{10.3},13.7,5.3,1.4,1.2,10.9,\textbf{11.9},–,2.7,3.2,1.0,\underline{2.2},1.4,6.5}}
        {{18.3,10.1,13.7,11.3,\textbf{7.9},\textbf{10.1},\textbf{12.6},7.8,–,\textbf{6.3},7.2,6.6,\underline{2.6},\underline{4.7},9.2}}
        {{18.1,10.1,\textbf{13.8},\textbf{12.8},7.8,9.9,11.4,7.6,–,5.5,\textbf{8.0},\textbf{6.7},\underline{\textbf{8.1}},\underline{\textbf{8.2}},\textbf{9.8}}}{\hlinep}

    \fewshotfloresresultstablerow{ar}
        {{25.5,18.7,25.7,18.9,15.6,17.8,23.8,13.1,13.3,–,15.4,19.4,1.8,0.9,16.1}}
        {{10.5,5.3,9.6,6.0,2.2,2.2,0.9,0.9,0.1,–,0.1,0.1,0.2,0.0,2.9}}
        {{\underline{\textbf{27.7}},\textbf{12.2},17.9,8.8,\textbf{8.5},9.1,\textbf{18.4},\textbf{8.9},0.8,–,7.7,7.8,\underline{3.4},\underline{3.7},10.4}}
        {{23.7,10.8,17.5,11.0,8.0,12.2,13.8,5.9,7.1,–,10.3,\textbf{8.0},\underline{8.0},\underline{8.0},11.1}}
        {{\underline{26.9},11.5,\textbf{19.8},\textbf{15.9},7.8,\textbf{14.5},13.6,6.3,\textbf{7.6},–,\textbf{11.0},\textbf{8.0},\underline{\textbf{8.8}},\underline{\textbf{9.3}},\textbf{12.4}}}{\hlinep}

    \fewshotfloresresultstablerow{sw}
        {{30.4,19.4,26.7,20.1,15.6,17.6,23.8,13.2,12.2,12.0,–,19.2,2.1,4.0,16.6}}
        {{5.0,2.9,3.9,2.8,1.7,1.8,1.3,1.3,0.5,0.5,–,0.4,0.1,0.1,1.7}}
        {{\underline{\textbf{31.6}},13.4,\textbf{21.8},15.4,\textbf{10.2},13.1,15.2,\textbf{9.5},\textbf{6.0},\textbf{8.9},–,\textbf{7.6},\underline{3.4},1.0,\textbf{12.1}}}
        {{27.0,12.6,19.0,15.1,9.2,12.2,\textbf{15.8},5.9,\textbf{6.0},8.3,–,6.5,\underline{\textbf{5.4}},\underline{6.0},11.5}}
        {{30.0,\textbf{13.5},20.0,\textbf{18.0},9.5,\textbf{14.5},\textbf{15.8},6.9,5.7,7.7,–,6.5,\underline{2.7},\underline{\textbf{7.0}},\textbf{12.1}}}{\hlinep}

    \fewshotfloresresultstablerow{hi}
        {{27.9,19.4,25.9,18.9,15.7,16.9,23.9,13.5,13.9,12.2,16.8,–,2.5,3.8,16.2}}
        {{1.2,0.9,1.4,0.8,0.4,0.4,0.3,0.2,0.1,0.1,0.1,–,0.1,0.2,0.5}}
        {{25.2,12.3,15.4,8.8,\textbf{9.8},11.5,11.3,\textbf{10.8},\textbf{8.5},6.1,4.7,–,1.5,1.9,9.8}}
        {{25.7,12.4,17.0,13.0,8.0,12.2,\textbf{15.4},7.2,4.4,7.4,\textbf{8.9},–,\underline{9.6},\underline{9.0},11.6}}
        {{\textbf{27.1},\textbf{12.6},\textbf{17.3},\textbf{14.3},9.0,\textbf{12.4},14.5,8.0,6.7,\textbf{8.1},\textbf{8.9},–,\underline{\textbf{10.2}},\underline{\textbf{12.8}},\textbf{12.5}}}{\hlinep}

    \fewshotfloresresultstablerow{my}
        {{10.0,6.9,10.4,8.5,6.0,6.7,9.5,5.7,6.1,4.6,7.2,9.1,–,2.5,7.2}}
        {{0.5,0.3,0.4,0.4,0.2,0.1,0.2,0.0,0.0,0.0,0.1,0.2,–,0.1,0.2}}
        {{\underline{14.1},\underline{7.6},10.1,3.8,5.7,\underline{7.1},8.9,\underline{\textbf{7.1}},\underline{\textbf{6.9}},3.6,3.5,\textbf{8.9},–,\underline{2.6},6.9}}
        {{\underline{\textbf{16.8}},\underline{8.5},\underline{\textbf{12.9}},\underline{11.0},\underline{6.7},6.1,9.2,5.2,2.9,\underline{\textbf{5.0}},\underline{\textbf{8.0}},7.0,–,\underline{5.7},\underline{8.1}}}
        {{\underline{16.4},\underline{\textbf{9.0}},\underline{11.9},\underline{\textbf{11.6}},\underline{\textbf{6.9}},\underline{\textbf{8.3}},\underline{\textbf{10.4}},5.5,3.6,\underline{4.8},6.4,7.1,–,\underline{\textbf{6.2}},\underline{\textbf{8.3}}}}{\hlinep}

    \fewshotfloresresultstablerow{ta}
        {{8.3,4.9,6.8,5.8,5.0,4.7,7.0,2.5,2.3,1.1,5.2,6.9,1.2,–,4.8}}
        {{1.0,0.5,0.8,0.5,0.2,0.3,0.3,0.1,0.2,0.1,0.1,0.2,0.0,–,0.3}}
        {{\underline{16.3},\underline{8.4},\underline{10.3},5.1,\underline{5.2},\underline{8.1},\underline{7.6},\underline{\textbf{8.1}},\underline{6.2},\underline{5.4},2.8,\underline{7.2},0.9,–,\underline{7.1}}}
        {{\underline{18.7},\underline{10.4},\underline{13.7},\underline{10.9},\underline{6.3},\underline{9.8},\underline{11.6},\underline{5.2},0.7,\underline{6.5},\underline{6.0},\underline{9.3},\underline{1.8},–,\underline{8.5}}}
        {{\underline{\textbf{20.4}},\underline{\textbf{10.5}},\underline{\textbf{14.7}},\underline{\textbf{12.9}},\underline{\textbf{8.1}},\underline{\textbf{10.6}},\underline{\textbf{13.2}},\underline{7.0},\underline{\textbf{6.8}},\underline{\textbf{6.6}},\underline{\textbf{8.3}},\underline{\textbf{10.1}},\underline{\textbf{2.6}},–,\underline{\textbf{10.1}}}}{\hlinep}

    \fewshotfloresresultstablerow{avg}
        {{26.0,20.2,26.7,20.0,16.7,18.5,24.5,14.1,13.5,11.8,16.3,19.3,2.1,2.5,16.6}}
        {{18.9,9.9,14.2,9.3,4.2,4.8,2.7,4.0,0.6,0.5,0.2,0.3,0.1,0.1,5.0}}
        {{\underline{\textbf{28.5}},14.9,20.6,14.4,\textbf{10.9},12.4,\textbf{18.5},\textbf{10.9},5.9,6.1,8.5,\textbf{9.7},\underline{5.8},\underline{3.5},12.2}}
        {{25.8,14.1,20.2,15.6,10.0,13.7,16.9,6.9,6.8,7.4,10.5,8.5,\underline{\textbf{8.1}},\underline{7.5},12.3}}
        {{\underline{27.1},\textbf{15.0},\textbf{20.9},\textbf{18.2},10.5,\textbf{14.8},17.1,7.9,\textbf{7.5},\textbf{8.0},\textbf{11.5},9.2,\underline{8.0},\underline{\textbf{9.7}},\textbf{13.2}}}{}
    \bottomrule[\heavyrulewidth]
    \end{tabular}
    \caption{Few-shot and unsupervised zero-shot machine translation results on FLORES-101 devtest (spBLEU). Source language in rows, target language in columns. $\text{GPT-3}_{\text{6.7B}}$ and $\text{XGLM}_{\text{7.5B}}$ use 32 examples from the dev set for few-shot learning. $\text{mT5}_{\text{3.7B}}$ uses 2 examples from the dev set for few-shot learning. Supervised results correspond to the M2M-124 615M model from \citet{flores101}. $\text{XGLM}_{\text{7.5B}}$ results correspond to the model from \citet{xglm}. Underline denotes better than supervised, bold denotes best of \text{GPT-3}, XGLM, and mT5. spBLEU computed using the implementation from \citet{flores101}.} 
    \label{table:fewshot-flores-results}
    \end{table}
}

\newcommand{\fewshotfloresresultstableabbrev}{
    \begin{table*}
    \centering
    \small
    \begin{tabular}{lrrrrr|r}
    \toprule[\heavyrulewidth]
                                                         && HR~$\rightarrow$~HR & LR~$\rightarrow$~HR & HR~$\rightarrow$~LR & LR~$\rightarrow$~LR & All   \\ \hline
    Number of Language Pairs & & 110 & 33 & 33 & 6 & 182 \\ \hline
    $\text{Supervised}$            &                                  & 21.5 & 10.3 & 8.6 & 4.3  & 16.6    \\ 
    $\text{GPT-3}_{\text{6.7B}}$ & (32\text{-shot})                   & 8.1 & 0.4 & 0.1 & 0.1  & 5.0    \\ 
    $\text{XGLM}_{\text{7.5B}}$ & (32\text{-shot})                    & 15.3 & 8.7 & 6.8 & 3.8  & 12.2             \\ 
    $\text{mT5}_{\text{3.7B}}\ + \  \text{\textsc{Sap}}$ & (2\text{-shot})     & 14.5 & 9.8 & 8.2 & 7.1  & 12.3             \\ 
    $\text{mT5}_{\text{3.7B}}\ + \  \text{\textsc{Sap}}$ & (zero\text{-shot})  & \textbf{15.5} & \textbf{10.7} & \textbf{9.1} & \textbf{8.2} & \textbf{13.2}            \\
    \bottomrule[\heavyrulewidth]
    \end{tabular}
    \caption{Abbreviated few-shot and unsupervised zero-shot machine translation results on FLORES-101 devtest (spBLEU). The matrix of full results can be found in Appendix \ref{sec:flores-results}. Results are average spBLEU scores over subsets of the 182 language pairs (\texttt{src}~$\rightarrow~$\texttt{tgt}) where ``LR'' is a low-resource language and ``HR'' is a high-resource language. ``All'' represents the average spBLEU score over all 182 language pairs. Supervised results correspond to the M2M-124 615M model from \citet{flores101}. $\text{XGLM}_{\text{7.5B}}$ results correspond to the model from \citet{xglm}. Bold denotes best of \text{GPT-3}, XGLM, and mT5. spBLEU computed using the implementation from \citet{flores101}.}
    \label{table:fewshot-flores-results-abbrev}
    \end{table*}
}

%% file: resources/zero-shot_results.tex
\newcommand{\selfamplificationtable}{
    \begin{table}[H]
    \centering
    \begin{tabularx}{\textwidth}{lXrr}
    \toprule[\heavyrulewidth]
                                                                                         && English-French  & French-English  \\ \hline
    $\text{GPT-3}_{\text{175B}}$   &(\text{self}\text{-amplified})  & \textbf{30.0}   & \textbf{31.8}    \\ 
    $\text{mT5}_{\text{3.7B}}$\ $+$ \ \text{\textsc{Sap}}  &(\text{self}\text{-amplified})       & 29.8            & 31.4             \\ 
    \bottomrule[\heavyrulewidth]
    \end{tabularx}
    \caption{Unsupervised zero-shot machine translation results on WMT14 English-French test set (SacreBLEU) \citep{wmt14,sacrebleu}. $\text{GPT-3}_{\text{175B}}\ \text{(self-amplified)}$ results correspond to the unsupervised zero-shot ``GPT-3 (self-amplified)'' results from \citet{gpt3unsupervised} prior to performing distillation, initial backtranslation, and iterative backtranslation which involved unsupervised weight updates. $\text{mT5}_{\text{3.7B}}\ \text{(self-amplified)}$ is our fully unsupervised zero-shot approach outlined in Section \ref{sec:zero-shot} with a 16 prompt ensemble. The SacreBLEU signature used also follows \citet{gpt3unsupervised}: \\{\small\texttt{BLEU+case.mixed+numrefs.1+smooth.exp+tok.intl+version.1.2.20})}}
    \label{table:selfamplification}
    \end{table}
}

%% file: resources/ablation_tables.tex
\newcommand{\firstwordablationtable}{
    \begin{table*}
    \centering
    \setlength{\tabcolsep}{2pt}
    \fontsize{8}{10}\selectfont
    \begin{tabularx}{\textwidth}{Xrr}
    \toprule[\heavyrulewidth]
                                                              & English-Russian & Russian-English   \\ \hline
    Prompting  ($\text{mT5}_\text{3.7B}$)\\{\tiny Using the full generation from the first time step only -- $G_0$}                                                    & 1.9             & 5.6               \\ \hline
    \textbf{S}equential \textbf{P}rompting ($\text{mT5}_\text{3.7B}$~+~\textsc{Sp}) \\{\tiny Concatenating the full generation at each time step  -- \texttt{CONCAT}($G_0$, \ldots, $G_{t-1}$)}                 & 9.3             & 17.9              \\ \hline
    \textbf{S}equential \textbf{A}utoregressive \textbf{P}rompting ($\text{mT5}_\text{3.7B}$~+~\textsc{Sap}) \\{\tiny Concatenating the first word of the generation at each time step -- \texttt{CONCAT}($F_0$, \ldots, $F_{t-1}$)} & \textbf{20.1}   & \textbf{26.9}     \\ 
    \bottomrule[\heavyrulewidth]
    \end{tabularx}
    \caption{Few-shot (2-shot) machine translation results on FLORES-101 devtest (spBLEU) using $\text{mT5}_{\text{3.7B}}$ as described in Section \ref{sec:few-shot}. In this experiment, over two language pairs, English-Russian and Russian-English, we compare a) simply prompting the model once and taking the full generation $G_0$ b) concatenating the full generation at each time step $G_t$ to the prompt in the next time step c) concatenating just the first word of the generation at each time step $F_t$ to the prompt in the next time step.}
    \label{table:firstword-ablation}
    \end{table*}
}

\newcommand{\mtscoreablationtable}{
    \begin{table}[H]
    \centering
    \begin{tabularx}{\textwidth}{Xrr}
    \toprule[\heavyrulewidth]
                                               & English-Russian & Russian-English \\ \hline
    Random Selection                           & 0.0             & 25.5            \\ 
    mT5Score Filtering and Selection           & \textbf{20.0}   & \textbf{26.3}   \\ 
    \bottomrule[\heavyrulewidth]
    \end{tabularx}
    \caption{Unsupervised zero-shot machine translation results on FLORES-101 devtest (spBLEU) using $\text{mT5}_{\text{3.7B}}$ as described in Section \ref{sec:zero-shot}. In this experiment, we ablate utilizing mT5Score to filter and select the high-quality synthetic examples during bootstrapping over two language pairs, English-Russian and Russian-English. When using random selection, the synthetic parallel examples choosen may be extremely low-quality or non-sensical leading to a 0.0 spBLEU score after self-amplification as shown for the English-Russian language pair. }
    \label{table:mtscore-ablation}
    \end{table}
}

\newcommand{\promptensembleablationtable}{
    \begin{table}[H]
    \centering
    \begin{tabularx}{\textwidth}{Xrr}
    \toprule[\heavyrulewidth]
                                               & English-Russian & Russian-English  \\ \hline
    Single Prompt                              & 20.0           & 26.3              \\
    4 Prompt Ensemble                          & \textbf{20.9}  & 27.9              \\ 
    8 Prompt Ensemble                          & 20.7           & \textbf{28.6}     \\ 
    16 Prompt Ensemble                         & \textbf{20.9}  & \textbf{28.6}     \\ 
    \bottomrule[\heavyrulewidth]
    \end{tabularx}
    \caption{Unsupervised zero-shot machine translation results on FLORES-101 devtest (spBLEU) using $\text{mT5}_{\text{3.7B}}$ as described in Section \ref{sec:zero-shot}. In this experiment, we compare utilizing a single few-shot prompt with two synthetic parallel examples to perform the final translation with utilizing an ensemble of 4, 8, and 16 distinct few-shot prompts each with two synthetic parallel examples that generate 4, 8, and 16 translations respectively from which the best translation (by mT5Score) is selected as the final translation over two language pairs, English-Russian and Russian-English.}
    \label{table:promptensemble-ablation}
    \end{table}
}

\newcommand{\englishcentricablationtable}{
    \begin{table}[H]
    \centering
    \begin{tabularx}{\textwidth}{Xrrr}
    \toprule[\heavyrulewidth]
                                & English-Russian & Russian-Chinese   \\ \hline
    Standard bootstrap          & 20.9            & 5.8              \\ 
    English-centric bootstrap   & \textbf{21.2}   & \textbf{8.3}     \\ 
    \bottomrule[\heavyrulewidth]
    \end{tabularx}
    \caption{Unsupervised zero-shot machine translation results on FLORES-101 devtest (spBLEU) using $\text{mT5}_{\text{3.7B}}$ as described in Section \ref{sec:zero-shot}. In this experiment, we ablate utilizing the English-centric bootstrap described in Section \ref{sec:englishcentric} over two language pairs, English-Russian and Russian-Chinese.}
    \label{table:englishcentric-ablation}
    \end{table}
}

\newcommand{\selfampablationtable}{
    \begin{table}[H]
    \centering
    \begin{tabularx}{\textwidth}{Xrr}
    \toprule[\heavyrulewidth]
                                               & English-Russian & Russian-English \\ \hline
    Standard                                   & 0.4             & 4.6            \\ 
    Bootstrapping and self-amplification       & \textbf{21.2}   & \textbf{27.9}   \\ 
    \bottomrule[\heavyrulewidth]
    \end{tabularx}
    \caption{Unsupervised zero-shot machine translation results on FLORES-101 devtest (spBLEU) using $\text{mT5}_{\text{3.7B}}$ as described in Section \ref{sec:zero-shot}. In this experiment, we compare the standard zero-shot performance of mT5 with \textsc{Sap} against the zero-shot performance achievable implementing the bootstrapping and self-amplification techniques from \citet{gpt3unsupervised} with the adaptations described in Section \ref{sec:zero-shot}. }
    \label{table:selfamp-ablation}
    \end{table}
}

%% file: resources/xquad_results.tex
\newcommand{\xquadtable}{
    \begin{table*}
    \fontsize{6}{9}\selectfont
    \centering
    \setlength{\tabcolsep}{2pt}
    \begin{tabular}{llccccccccccc|c}
    \toprule[\heavyrulewidth]
                                                                         && en      & ar      & de      & el      & es      & hi      & ru      & th      & tr      & vi      & zh      & avg  \\ \hline
    $\text{XGLM}_{\text{7.5B}}$ & (\text{zero-shot})                      & 19.5/31.9 & 12.9/29.6 & 12.2/25.3 & 7.2/28.2 & 12.5/24.0 & \textbf{11.0}/14.0 & 10.9/27.8 & \textbf{16.8}/\textbf{26.4} & 13.6/26.8 & 12.5/21.2 & 13.2/20.3 & 12.9/25.0 \\
    $\text{mT5}_{\text{3.7B}}$\ $+$ \ $\text{\textsc{Sap}}$ & (\text{zero-shot})   & \textbf{25.0}/\textbf{48.8} & \textbf{17.4}/\textbf{39.4} & \textbf{19.4}/\textbf{43.0} & \textbf{9.7}/\textbf{41.0} & \textbf{15.0}/\textbf{42.1} & 6.6/\textbf{32.1} & \textbf{16.1}/\textbf{39.0} & 2.8/17.4 & \textbf{15.8}/\textbf{37.0} & \textbf{18.2}/\textbf{41.9} & \textbf{15.0}/\textbf{29.0} & \textbf{14.6}/\textbf{37.3} \\
    \bottomrule[\heavyrulewidth]
    \end{tabular}
    \caption{Zero-shot multilingual question answering results (EM/F1) on the XQuAD test set \citep{xquad}.}
    \label{table:xquad}
    \end{table*}
}

%% file: resources/squad_and_summarization_results.tex
\newcommand{\squadandsummarizationtable}{
    \begin{table}[t]
    \centering
    \captionsetup{justification=centering}
    \setlength{\tabcolsep}{3pt}
    \small

    \begin{minipage}[t]{0.42\linewidth}\centering
        \fontsize{8}{9}\selectfont
        \begin{tabular}{llrrr}
            \toprule[\heavyrulewidth]
                                                                                 && EM              & F1             \\ \hline
            \textit{Zero-shot}                                                                                       \\ \hline 
            $\text{T5+LM}_{\text{3B}}$                     & (zero\text{-shot})   & 23.5            & 48.4           \\ 
            $\text{mT5}_{\text{3.7B}}$ $+$ \ $\text{\textsc{Sap}}$  & (zero\text{-shot})   & \textbf{30.2}    & \textbf{54.0} \\ \hline
            \textit{Few-shot}                                                                                        \\ \hline   
            $\text{mT5}_{\text{3.7B}}$\                    & (16\text{-shot})     & 23.0            & 54.5            \\ 
            $\text{mT5}_{\text{3.7B}}$\ $+$ \ $\text{\textsc{Sap}}$ & (16\text{-shot})     & \textbf{35.4}   & \textbf{60.0}   \\ 
            \bottomrule[\heavyrulewidth]
        \end{tabular}
        \caption{Zero-shot and few-shot question answering results on the SQuAD v1.1 dev set \citep{squad}.}
        \label{table:squad}
    \end{minipage}\hfill%
    \begin{minipage}[t]{0.57\linewidth}\centering
        \fontsize{8}{9}\selectfont
        \begin{tabular}{llrrr}
        \toprule[\heavyrulewidth]
                                                                          && ROUGE-1         & ROUGE-2        & ROUGE-L      \\ \hline
        \textit{Zero-shot}                                                                                       \\ \hline 
        $\text{T5+LM}_{\text{3B}}$ & (zero\text{-shot})                       &  5.3            &  0.6            &  4.9         \\ 
        $\text{mT5}_{\text{3.7B}}$ & (zero\text{-shot})                       & 15.4            & 4.6            & 14.5         \\ 
        $\text{mT5}_{\text{3.7B}}$\ $+$ \ $\text{\textsc{Sap}}$ & (zero\text{-shot})   & \textbf{22.0}   & \textbf{7.4}   & \textbf{20.1} \\ \hline
        \textit{Few-shot}                                                                                       \\ \hline 
        $\text{T5+LM}_{\text{3B}}$ & (2\text{-shot})                       & 14.1            & 4.4            & 13.2         \\ 
        $\text{mT5}_{\text{3.7B}}$ & (2\text{-shot})                       & 15.9            & 4.5            & 15.0         \\ 
        $\text{mT5}_{\text{3.7B}}$\ $+$ \ $\text{\textsc{Sap}}$ & (2\text{-shot})   & \textbf{22.0}   & \textbf{6.8}   & \textbf{20.3} \\ 
        \bottomrule[\heavyrulewidth]
        \end{tabular}
        \caption{Zero-shot and few-shot summarization results on the CNN / Daily Mail v3.0.0 test set evaluated with ROUGE \citep{cnndailymail,cnndailymail2,cnndailymail3,rouge}.}
        \label{table:summarization}
    \end{minipage}
    \end{table}
}

%% file: resources/selected_generations.tex
\newcommand{\exampleMT}[9]{
\vspace{1em}
\noindent\underline{Task:} #1-shot Machine Translation (Example \##2) \\
\\ \underline{Dataset:} FLORES-101 (#3~$\rightarrow$~#4) \\
\\ \underline{Prompt Template:} \\ \\
\texttt{\small{Translate #3 to #4.\\
\ifthenelse{\equal{#1}{Few}}{{\{\{examples\}\}\\}}{{}}
#3: \{\{source\_text\}\}\\
#4: }}
\\ \\ \underline{Ground Truth:} \\ \\
\texttt{\small{#6}}
\\ \\ \underline{Generation ($\text{mT5}_\text{3.7B}$~+~\textsc{Sap}):} \\ \\
\texttt{\small{#7}}
\\ \\ \underline{Generation ($\text{mT5}_\text{3.7B}$):} \\ \\
\texttt{\small{#8}}
\\ \\ \underline{Commentary:} \\ \\
#9
\raggedbottom
\pagebreak
}

\newcommand{\exampleQA}[9]{
\vspace{1em}
\noindent\underline{Task:} #1-shot Question Answering (Example \##2) \\
\\ \underline{Dataset:} #3 \\
\\ \underline{Prompt Template:} \\ \\
\texttt{\small{Answer the question based on the following passage.\\\\
\ifthenelse{\equal{#1}{Few}}{{\{\{examples\}\}\\\\}}{{}}
Passage: \{\{passage\}\}\\
Question: \{\{question\}\}\\
Answer: }}
\\ \\ \underline{Passage:} \\ \\
\texttt{\small{#4}}
\\ \\ \underline{Question:} \\ \\
\texttt{\small{#5}}
\\ \\ \underline{Ground Truth:} \\ \\
\texttt{\small{#6}}
\\ \\ \underline{Generation ($\text{mT5}_\text{3.7B}$~+~\textsc{Sap}):} \\ \\
\texttt{\small{#7}}
\\ \\ \underline{Generation ($\text{mT5}_\text{3.7B}$):} \\ \\
\texttt{\small{#8}}
\\ \\ \underline{Commentary:} \\ \\
#9
\raggedbottom
\pagebreak
}

\newcommand{\exampleSUM}[6]{
\vspace{1em}
\noindent\underline{Task:} Few-shot Summarization (Example \##1) \\
\\ \underline{Dataset:} CNN / Daily Mail v3.0.0 \\
\\ \underline{Prompt Template:} \\ \\
\texttt{\small{Summarize the article.\\\\
\{\{examples\}\}\\\\
Article: \{\{article\}\}\\
Summary: }}
\\ \\ \underline{Article:} \\ \\
\texttt{\small{#2}}
\\ \\ \underline{Summary:} \\ \\
\texttt{\small{#3}}
\\ \\ \underline{Generation ($\text{mT5}_\text{3.7B}$~+~\textsc{Sap}):} \\ \\
\texttt{\small{#4}}
\\ \\ \underline{Generation ($\text{mT5}_\text{3.7B}$):} \\ \\
\texttt{\small{#5}}
\\ \\ \underline{Commentary:} \\ \\
#6
\raggedbottom
\pagebreak
}

\newcommand{\selectedgenerations} {

\exampleMT{Few}
{1}
{Arabic}
{English}
{
}
{The 802.11n standard operates on both the 2.4Ghz and 5.0Ghz frequencies.}
{The wireless standard 802.11n operates at the frequency of 2.4 GHz and 5 GHz.}
{The 802.11n wireless standard operates at 2.4 and 5.0}
{\textsc{Sap} generates a lengthier and more fluent translation and correctly translates the units of the frequencies. Both generations add the word ``wireless'' which is used correctly and is likely to appear next to the words ``802.11n'' and ``standard'' but does not exist in the ground truth translation.}

\exampleMT{Few}
{2}
{Russian}
{English}
{}
{In 1956 Słania moved to Sweden, where three years later he began work for the Swedish Post Office and became their chief engraver.}
{In 1956, Slania moved to Sweden, where he worked for three years for the Swedish Post Office and became its chief engraved worker.}
{In 1956, Slanya moved to Sweden and became...}
{\textsc{Sap} generates a full length translation and more correctly translates ``Słania'' to ``Slania'' instead of ``Slanya''. While the translation without \textsc{Sap} only generates a partial translation, the word ``became'' indicates the direction of translation would be less close to the ground truth translation than the direction of translation taken by \textsc{Sap}. Notably, \textsc{Sap} produces a relatively high-quality translation, but a common failure mode is displayed in this example. \textsc{Sap} translates ``chief  engraver'' to ``chief engraved worker'' which is an imperfect paraphrase likely due to an imperfect multilingual alignment of the word  ``engraver'' in the embedding space of the model.}

\exampleQA{Few}
{1}
{SQuAD v1.1}
{In 1874, Tesla evaded being drafted into the Austro-Hungarian Army in Smiljan by running away to Tomingaj, near Gračac. There, he explored the mountains in hunter's garb. Tesla said that this contact with nature made him stronger, both physically and mentally. He read many books while in Tomingaj, and later said that Mark Twain's works had helped him to miraculously recover from his earlier illness.}
{Why did Tesla avoid by fleeing Smiljan?}
{being drafted into the Austro-Hungarian Army}
{because he was ill and wanted to avoid being drafted into the Austro-Hungarian Army}
{because he was ill and could not leave the country}
{In this example, the grammaticality of the question itself (``\textit{Why} did Tesla avoid by fleeing Smiljan?'' vs. ``\textit{What} did Tesla avoid by fleeing Smiljan?'') has issues. This seems to cause both generations to attempt to answer a ``why'' style question with ``because'' instead of a ``what'' style question. Notably, the answer generated by \textsc{Sap} does eventually reach correct answer where as the the answer generated without \textsc{Sap} hallucinates a fact: ``he [Tesla] \ldots could not leave the country''.}

\exampleQA{Few}
{2}
{SQuAD v1.1}
{The Broncos took an early lead in Super Bowl 50 and never trailed. Newton was limited by Denver's defense, which sacked him seven times and forced him into three turnovers, including a fumble which they recovered for a touchdown. Denver linebacker Von Miller was named Super Bowl MVP, recording five solo tackles, 2½ sacks, and two forced fumbles.}
{Who won the MVP for the Super Bowl?}
{Von Miller}
{Von Miller}
{Von Miller, Denver, NFL, NFLPA}
{Without using \textsc{Sap}, the answer generated is correct, but contains hallucinated tokens towards the end of the generation.}

\exampleSUM{1}
{London (CNN)The Hatton Garden heist, as it will surely come to be known, was every safe deposit box holder's nightmare, every movie director's dream. Thieves using heavy cutting equipment and rappelling gear broke into the vault of an esteemed 60-year-old safe deposit company in the heart of London over the past holiday weekend, possibly taking advantage of as many as four days to rifle through an uncounted number of safe deposit boxes. And they reportedly got away with hundreds of thousands of pounds worth of gems and cash -- even, in the educated guess of one former police official, as much as 200 million pounds, or\ldots[truncated]}
{Robbers may have taken advantage of a four-day holiday weekend. Estimates of the value of the items taken rage from hundreds of thousands of pounds to 200 million pounds. The heist took place in a historic heart of London's jewelry business.}
{The Hatton Garden heist was a nightmare for jewelers and their customers. They were 'treated like criminals', while officers said they would double bail for anyone who was arrested before. They were arrested while on the sidewalk .}
{The latest jewel heist in London took place over the weekend.}
{\textsc{Sap} generates a lengthier generation with multiple sentences summarizing the article. Some of the facts presented in the generation, however, are not coherent with the article.}

\exampleSUM{2}
{Sanaa, Yemen (CNN)Al Qaeda fighters attacked a prison in the coastal Yemeni city of Al Mukallah early Thursday, freeing at least 270 prisoners, a third of whom have al Qaeda links, a senior Defense Ministry official has told CNN. Khaled Batarfi, a senior al Qaeda figure, was among the escapees, officials said. Dozens of attackers took control of government buildings, including the city's Central Prison, Central Bank and radio station during the assault early Thursday, according to officials. Government troops arrived early Thursday and clashed with the al Qaeda fighters, and most of the militants fled,\ldots[truncated]}
{Al Qaeda fighters attack a prison and other government buildings, freeing many prisoners. Government troops clash with the fighters, most of whom flee. Yemen is descending into chaos as a Shia-Sunni conflict draws in regional rivals Saudi Arabia and Iran.}
{Yemen's al Qaeda fighters escape from prison after clash with government forces.}
{Yemen's prisoners escape from al Qaeda fighting in Yemen}
{\textsc{Sap} correctly characterizes the major theme of the article in its summary. Without \textsc{Sap}, the generation devolves in to an incoherent fact (``prisoners escape from al Qaeda fighting'') and only a partial summary is generated. The ground truth summary, however, is notably longer and contains multiple sentences, while the summary generated by \textsc{Sap} in this instance is only a single sentence.}

}

%% file: iclr2023_conference.bbl
\begin{thebibliography}{47}
\providecommand{\natexlab}[1]{#1}
\providecommand{\url}[1]{\texttt{#1}}
\expandafter\ifx\csname urlstyle\endcsname\relax
  \providecommand{\doi}[1]{doi: #1}\else
  \providecommand{\doi}{doi: \begingroup \urlstyle{rm}\Url}\fi

\bibitem[Andonian et~al.(2021)Andonian, Biderman, Black, Gali, Gao, Hallahan,
  Levy-Kramer, Leahy, Nestler, Parker, Pieler, Purohit, Songz, Phil, and
  Weinbach]{gpt-neo-x}
Alex Andonian, Stella Biderman, Sid Black, Preetham Gali, Leo Gao, Eric
  Hallahan, Josh Levy-Kramer, Connor Leahy, Lucas Nestler, Kip Parker, Michael
  Pieler, Shivanshu Purohit, Tri Songz, Wang Phil, and Samuel Weinbach.
\newblock {GPT-NeoX: Large Scale Autoregressive Language Modeling in PyTorch},
  8 2021.
\newblock URL \url{https://www.github.com/eleutherai/gpt-neox}.

\bibitem[Artetxe et~al.(2020)Artetxe, Ruder, and Yogatama]{xquad}
Mikel Artetxe, Sebastian Ruder, and Dani Yogatama.
\newblock On the cross-lingual transferability of monolingual representations.
\newblock In \emph{Proceedings of the 58th Annual Meeting of the Association
  for Computational Linguistics}, pp.\  4623--4637, 2020.

\bibitem[BigScience(2022)]{bloom}
BigScience.
\newblock {BigScience Language Open-science Open-access Multilingual (BLOOM)
  Language Model}.
\newblock \url{https://huggingface.co/bigscience/bloom-7b1}, May 2022.

\bibitem[Black et~al.(2021)Black, Gao, Wang, Leahy, and Biderman]{gpt-neo}
Sid Black, Leo Gao, Phil Wang, Connor Leahy, and Stella Biderman.
\newblock {GPT-Neo: Large Scale Autoregressive Language Modeling with
  Mesh-Tensorflow}, March 2021.
\newblock URL \url{https://doi.org/10.5281/zenodo.5297715}.

\bibitem[Bojar et~al.(2014)Bojar, Buck, Federmann, Haddow, Koehn, Leveling,
  Monz, Pecina, Post, Saint-Amand, Soricut, Specia, and Tamchyna]{wmt14}
Ondrej Bojar, Christian Buck, Christian Federmann, Barry Haddow, Philipp Koehn,
  Johannes Leveling, Christof Monz, Pavel Pecina, Matt Post, Herve Saint-Amand,
  Radu Soricut, Lucia Specia, and Ale\v{s} Tamchyna.
\newblock Findings of the 2014 workshop on statistical machine translation.
\newblock In \emph{Proceedings of the Ninth Workshop on Statistical Machine
  Translation}, pp.\  12--58, Baltimore, Maryland, USA, June 2014. Association
  for Computational Linguistics.
\newblock URL \url{http://www.aclweb.org/anthology/W/W14/W14-3302}.

\bibitem[Brown et~al.(2020)Brown, Mann, Ryder, Subbiah, Kaplan, Dhariwal,
  Neelakantan, Shyam, Sastry, Askell, Agarwal, Herbert-Voss, Krueger, Henighan,
  Child, Ramesh, Ziegler, Wu, Winter, Hesse, Chen, Sigler, Litwin, Gray, Chess,
  Clark, Berner, McCandlish, Radford, Sutskever, and Amodei]{gpt3}
Tom Brown, Benjamin Mann, Nick Ryder, Melanie Subbiah, Jared~D Kaplan, Prafulla
  Dhariwal, Arvind Neelakantan, Pranav Shyam, Girish Sastry, Amanda Askell,
  Sandhini Agarwal, Ariel Herbert-Voss, Gretchen Krueger, Tom Henighan, Rewon
  Child, Aditya Ramesh, Daniel Ziegler, Jeffrey Wu, Clemens Winter, Chris
  Hesse, Mark Chen, Eric Sigler, Mateusz Litwin, Scott Gray, Benjamin Chess,
  Jack Clark, Christopher Berner, Sam McCandlish, Alec Radford, Ilya Sutskever,
  and Dario Amodei.
\newblock Language models are few-shot learners.
\newblock In H.~Larochelle, M.~Ranzato, R.~Hadsell, M.~F. Balcan, and H.~Lin
  (eds.), \emph{Advances in Neural Information Processing Systems}, volume~33,
  pp.\  1877--1901. Curran Associates, Inc., 2020.
\newblock URL
  \url{https://proceedings.neurips.cc/paper/2020/file/1457c0d6bfcb4967418bfb8ac142f64a-Paper.pdf}.

\bibitem[Clark et~al.(2020)Clark, Luong, Le, and Manning]{electra}
Kevin Clark, Minh-Thang Luong, Quoc~V Le, and Christopher~D Manning.
\newblock Electra: Pre-training text encoders as discriminators rather than
  generators.
\newblock \emph{arXiv preprint arXiv:2003.10555}, 2020.

\bibitem[Conneau et~al.(2020)Conneau, Khandelwal, Goyal, Chaudhary, Wenzek,
  Guzm{\'a}n, Grave, Ott, Zettlemoyer, and Stoyanov]{xlm}
Alexis Conneau, Kartikay Khandelwal, Naman Goyal, Vishrav Chaudhary, Guillaume
  Wenzek, Francisco Guzm{\'a}n, Edouard Grave, Myle Ott, Luke Zettlemoyer, and
  Veselin Stoyanov.
\newblock Unsupervised cross-lingual representation learning at scale.
\newblock In \emph{ACL}, 2020.

\bibitem[Devlin et~al.(2019)Devlin, Chang, Lee, and Toutanova]{bert}
Jacob Devlin, Ming-Wei Chang, Kenton Lee, and Kristina Toutanova.
\newblock Bert: Pre-training of deep bidirectional transformers for language
  understanding.
\newblock In \emph{Proceedings of the 2019 Conference of the North American
  Chapter of the Association for Computational Linguistics: Human Language
  Technologies, Volume 1 (Long and Short Papers)}, pp.\  4171--4186, 2019.

\bibitem[Goyal et~al.(2021)Goyal, Gao, Chaudhary, Chen, Wenzek, Ju, Krishnan,
  Ranzato, Guzman, and Fan]{flores101}
Naman Goyal, Cynthia Gao, Vishrav Chaudhary, Peng-Jen Chen, Guillaume Wenzek,
  Da~Ju, Sanjana Krishnan, Marc'Aurelio Ranzato, Francisco Guzman, and Angela
  Fan.
\newblock The flores-101 evaluation benchmark for low-resource and multilingual
  machine translation.
\newblock \emph{arXiv preprint arXiv:2106.03193}, 2021.

\bibitem[Han et~al.(2021)Han, Babuschkin, Edwards, Neelakantan, Xu, Polu, Ray,
  Shyam, Ramesh, Radford, et~al.]{gpt3unsupervised}
Jesse~Michael Han, Igor Babuschkin, Harrison Edwards, Arvind Neelakantan, Tao
  Xu, Stanislas Polu, Alex Ray, Pranav Shyam, Aditya Ramesh, Alec Radford,
  et~al.
\newblock Unsupervised neural machine translation with generative language
  models only.
\newblock \emph{arXiv preprint arXiv:2110.05448}, 2021.

\bibitem[Hermann et~al.(2015)Hermann, Kociský, Grefenstette, Espeholt, Kay,
  Suleyman, and Blunsom]{cnndailymail3}
Karl~Moritz Hermann, Tomás Kociský, Edward Grefenstette, Lasse Espeholt, Will
  Kay, Mustafa Suleyman, and Phil Blunsom.
\newblock Teaching machines to read and comprehend.
\newblock In \emph{NIPS}, pp.\  1693--1701, 2015.
\newblock URL
  \url{http://papers.nips.cc/paper/5945-teaching-machines-to-read-and-comprehend}.

\bibitem[Kaplan et~al.(2020)Kaplan, McCandlish, Henighan, Brown, Chess, Child,
  Gray, Radford, Wu, and Amodei]{scaling}
Jared Kaplan, Sam McCandlish, Tom Henighan, Tom~B Brown, Benjamin Chess, Rewon
  Child, Scott Gray, Alec Radford, Jeffrey Wu, and Dario Amodei.
\newblock Scaling laws for neural language models.
\newblock \emph{arXiv preprint arXiv:2001.08361}, 2020.

\bibitem[Lan et~al.(2019)Lan, Chen, Goodman, Gimpel, Sharma, and
  Soricut]{albert}
Zhenzhong Lan, Mingda Chen, Sebastian Goodman, Kevin Gimpel, Piyush Sharma, and
  Radu Soricut.
\newblock {ALBERT:} {A} lite {BERT} for self-supervised learning of language
  representations.
\newblock \emph{CoRR}, abs/1909.11942, 2019.
\newblock URL \url{http://arxiv.org/abs/1909.11942}.

\bibitem[Lester et~al.(2021)Lester, Al-Rfou, and
  Constant]{lester-etal-2021-power}
Brian Lester, Rami Al-Rfou, and Noah Constant.
\newblock The power of scale for parameter-efficient prompt tuning.
\newblock In \emph{Proceedings of the 2021 Conference on Empirical Methods in
  Natural Language Processing}, pp.\  3045--3059, Online and Punta Cana,
  Dominican Republic, November 2021. Association for Computational Linguistics.
\newblock \doi{10.18653/v1/2021.emnlp-main.243}.
\newblock URL \url{https://aclanthology.org/2021.emnlp-main.243}.

\bibitem[Lewis et~al.(2019)Lewis, Liu, Goyal, Ghazvininejad, Mohamed, Levy,
  Stoyanov, and Zettlemoyer]{bart}
Mike Lewis, Yinhan Liu, Naman Goyal, Marjan Ghazvininejad, Abdelrahman Mohamed,
  Omer Levy, Ves Stoyanov, and Luke Zettlemoyer.
\newblock Bart: Denoising sequence-to-sequence pre-training for natural
  language generation, translation, and comprehension.
\newblock \emph{arXiv preprint arXiv:1910.13461}, 2019.

\bibitem[Lhoest et~al.(2021)Lhoest, Villanova~del Moral, Jernite, Thakur, von
  Platen, Patil, Chaumond, Drame, Plu, Tunstall, Davison, {\v{S}}a{\v{s}}ko,
  Chhablani, Malik, Brandeis, Le~Scao, Sanh, Xu, Patry, McMillan-Major, Schmid,
  Gugger, Delangue, Matussi{\`e}re, Debut, Bekman, Cistac, Goehringer, Mustar,
  Lagunas, Rush, and Wolf]{lhoest-etal-2021-datasets}
Quentin Lhoest, Albert Villanova~del Moral, Yacine Jernite, Abhishek Thakur,
  Patrick von Platen, Suraj Patil, Julien Chaumond, Mariama Drame, Julien Plu,
  Lewis Tunstall, Joe Davison, Mario {\v{S}}a{\v{s}}ko, Gunjan Chhablani,
  Bhavitvya Malik, Simon Brandeis, Teven Le~Scao, Victor Sanh, Canwen Xu,
  Nicolas Patry, Angelina McMillan-Major, Philipp Schmid, Sylvain Gugger,
  Cl{\'e}ment Delangue, Th{\'e}o Matussi{\`e}re, Lysandre Debut, Stas Bekman,
  Pierric Cistac, Thibault Goehringer, Victor Mustar, Fran{\c{c}}ois Lagunas,
  Alexander Rush, and Thomas Wolf.
\newblock Datasets: A community library for natural language processing.
\newblock In \emph{Proceedings of the 2021 Conference on Empirical Methods in
  Natural Language Processing: System Demonstrations}, pp.\  175--184, Online
  and Punta Cana, Dominican Republic, November 2021. Association for
  Computational Linguistics.
\newblock \doi{10.18653/v1/2021.emnlp-demo.21}.
\newblock URL \url{https://aclanthology.org/2021.emnlp-demo.21}.

\bibitem[Li \& Liang(2021)Li and Liang]{prefixtuning}
Xiang~Lisa Li and Percy Liang.
\newblock Prefix-tuning: Optimizing continuous prompts for generation.
\newblock In \emph{Proceedings of the 59th Annual Meeting of the Association
  for Computational Linguistics and the 11th International Joint Conference on
  Natural Language Processing (Volume 1: Long Papers)}, pp.\  4582--4597, 2021.

\bibitem[Lin(2004)]{rouge}
Chin-Yew Lin.
\newblock {ROUGE}: A package for automatic evaluation of summaries.
\newblock In \emph{Text Summarization Branches Out}, pp.\  74--81, Barcelona,
  Spain, July 2004. Association for Computational Linguistics.
\newblock URL \url{https://aclanthology.org/W04-1013}.

\bibitem[Lin et~al.(2021)Lin, Mihaylov, Artetxe, Wang, Chen, Simig, Ott, Goyal,
  Bhosale, Du, Pasunuru, Shleifer, Koura, Chaudhary, O'Horo, Wang, Zettlemoyer,
  Kozareva, Diab, Stoyanov, and Li]{xglm}
Xi~Victoria Lin, Todor Mihaylov, Mikel Artetxe, Tianlu Wang, Shuohui Chen,
  Daniel Simig, Myle Ott, Naman Goyal, Shruti Bhosale, Jingfei Du, Ramakanth
  Pasunuru, Sam Shleifer, Punit~Singh Koura, Vishrav Chaudhary, Brian O'Horo,
  Jeff Wang, Luke Zettlemoyer, Zornitsa Kozareva, Mona~T. Diab, Veselin
  Stoyanov, and Xian Li.
\newblock Few-shot learning with multilingual language models.
\newblock \emph{CoRR}, abs/2112.10668, 2021.
\newblock URL \url{https://arxiv.org/abs/2112.10668}.

\bibitem[Liu et~al.(2022)Liu, Tam, Muqeeth, Mohta, Huang, Bansal, and
  Raffel]{parameff}
Haokun Liu, Derek Tam, Mohammed Muqeeth, Jay Mohta, Tenghao Huang, Mohit
  Bansal, and Colin Raffel.
\newblock Few-shot parameter-efficient fine-tuning is better and cheaper than
  in-context learning.
\newblock \emph{arXiv preprint arXiv:2205.05638}, 2022.

\bibitem[Liu et~al.(2021)Liu, Yuan, Fu, Jiang, Hayashi, and
  Neubig]{promptingsurvey}
Pengfei Liu, Weizhe Yuan, Jinlan Fu, Zhengbao Jiang, Hiroaki Hayashi, and
  Graham Neubig.
\newblock Pre-train, prompt, and predict: A systematic survey of prompting
  methods in natural language processing.
\newblock \emph{arXiv preprint arXiv:2107.13586}, 2021.

\bibitem[Liu et~al.(2019)Liu, Ott, Goyal, Du, Joshi, Chen, Levy, Lewis,
  Zettlemoyer, and Stoyanov]{roberta}
Yinhan Liu, Myle Ott, Naman Goyal, Jingfei Du, Mandar Joshi, Danqi Chen, Omer
  Levy, Mike Lewis, Luke Zettlemoyer, and Veselin Stoyanov.
\newblock Roberta: A robustly optimized bert pretraining approach.
\newblock \emph{arXiv preprint arXiv:1907.11692}, 2019.

\bibitem[Liu et~al.(2020)Liu, Gu, Goyal, Li, Edunov, Ghazvininejad, Lewis, and
  Zettlemoyer]{mbart}
Yinhan Liu, Jiatao Gu, Naman Goyal, Xian Li, Sergey Edunov, Marjan
  Ghazvininejad, Mike Lewis, and Luke Zettlemoyer.
\newblock Multilingual denoising pre-training for neural machine translation.
\newblock \emph{Transactions of the Association for Computational Linguistics},
  8:\penalty0 726--742, 2020.

\bibitem[Min et~al.(2021)Min, Lewis, Zettlemoyer, and Hajishirzi]{metalicl}
Sewon Min, Mike Lewis, Luke Zettlemoyer, and Hannaneh Hajishirzi.
\newblock Metaicl: Learning to learn in context.
\newblock \emph{arXiv preprint arXiv:2110.15943}, 2021.

\bibitem[Nallapati et~al.(2016)Nallapati, Zhou, dos Santos, G\.{u}l{\c{c}}ehre,
  and Xiang]{cnndailymail}
Ramesh Nallapati, Bowen Zhou, Cicero dos Santos, {\c{C}}a{\u{g}}lar
  G\.{u}l{\c{c}}ehre, and Bing Xiang.
\newblock Abstractive text summarization using sequence-to-sequence {RNN}s and
  beyond.
\newblock In \emph{Proceedings of The 20th {SIGNLL} Conference on Computational
  Natural Language Learning}, pp.\  280--290, Berlin, Germany, August 2016.
  Association for Computational Linguistics.
\newblock \doi{10.18653/v1/K16-1028}.
\newblock URL \url{https://aclanthology.org/K16-1028}.

\bibitem[Ni \& Kao(2022)Ni and Kao]{electrazero}
Shiwen Ni and Hung-Yu Kao.
\newblock Electra is a zero-shot learner, too.
\newblock \emph{arXiv preprint arXiv:2207.08141}, 2022.

\bibitem[Ouyang et~al.(2022)Ouyang, Wu, Jiang, Almeida, Wainwright, Mishkin,
  Zhang, Agarwal, Slama, Ray, et~al.]{instructgpt}
Long Ouyang, Jeff Wu, Xu~Jiang, Diogo Almeida, Carroll~L Wainwright, Pamela
  Mishkin, Chong Zhang, Sandhini Agarwal, Katarina Slama, Alex Ray, et~al.
\newblock Training language models to follow instructions with human feedback.
\newblock \emph{arXiv preprint arXiv:2203.02155}, 2022.

\bibitem[Post(2018)]{sacrebleu}
Matt Post.
\newblock A call for clarity in reporting {BLEU} scores.
\newblock In \emph{Proceedings of the Third Conference on Machine Translation:
  Research Papers}, pp.\  186--191, Brussels, Belgium, October 2018.
  Association for Computational Linguistics.
\newblock \doi{10.18653/v1/W18-6319}.
\newblock URL \url{https://aclanthology.org/W18-6319}.

\bibitem[Radford et~al.(2019)Radford, Wu, Child, Luan, Amodei, and
  Sutskever]{gpt2}
Alec Radford, Jeff Wu, Rewon Child, David Luan, Dario Amodei, and Ilya
  Sutskever.
\newblock Language models are unsupervised multitask learners.
\newblock 2019.

\bibitem[Raffel et~al.(2020)Raffel, Shazeer, Roberts, Lee, Narang, Matena,
  Zhou, Li, and Liu]{t5}
Colin Raffel, Noam Shazeer, Adam Roberts, Katherine Lee, Sharan Narang, Michael
  Matena, Yanqi Zhou, Wei Li, and Peter~J Liu.
\newblock Exploring the limits of transfer learning with a unified text-to-text
  transformer.
\newblock \emph{Journal of Machine Learning Research}, 21:\penalty0 1--67,
  2020.

\bibitem[Rajpurkar et~al.(2016)Rajpurkar, Zhang, Lopyrev, and Liang]{squad}
Pranav Rajpurkar, Jian Zhang, Konstantin Lopyrev, and Percy Liang.
\newblock Squad: 100,000+ questions for machine comprehension of text.
\newblock In \emph{Proceedings of the 2016 Conference on Empirical Methods in
  Natural Language Processing}, pp.\  2383--2392, 2016.

\bibitem[Sanh et~al.(2022)Sanh, Webson, Raffel, Bach, Sutawika, Alyafeai,
  Chaffin, Stiegler, Le~Scao, Raja, et~al.]{t0}
Victor Sanh, Albert Webson, Colin Raffel, Stephen Bach, Lintang Sutawika, Zaid
  Alyafeai, Antoine Chaffin, Arnaud Stiegler, Teven Le~Scao, Arun Raja, et~al.
\newblock Multitask prompted training enables zero-shot task generalization.
\newblock In \emph{The Tenth International Conference on Learning
  Representations}, 2022.

\bibitem[Schick \& Sch{\"u}tze(2021{\natexlab{a}})Schick and
  Sch{\"u}tze]{cloze}
Timo Schick and Hinrich Sch{\"u}tze.
\newblock Exploiting cloze-questions for few-shot text classification and
  natural language inference.
\newblock In \emph{EACL}, 2021{\natexlab{a}}.

\bibitem[Schick \& Sch{\"u}tze(2021{\natexlab{b}})Schick and
  Sch{\"u}tze]{cloze2}
Timo Schick and Hinrich Sch{\"u}tze.
\newblock It’s not just size that matters: Small language models are also
  few-shot learners.
\newblock In \emph{Proceedings of the 2021 Conference of the North American
  Chapter of the Association for Computational Linguistics: Human Language
  Technologies}, pp.\  2339--2352, 2021{\natexlab{b}}.

\bibitem[See et~al.(2017)See, Liu, and Manning]{cnndailymail2}
Abigail See, Peter~J. Liu, and Christopher~D. Manning.
\newblock Get to the point: Summarization with pointer-generator networks.
\newblock In \emph{Proceedings of the 55th Annual Meeting of the Association
  for Computational Linguistics (Volume 1: Long Papers)}, pp.\  1073--1083,
  Vancouver, Canada, July 2017. Association for Computational Linguistics.
\newblock \doi{10.18653/v1/P17-1099}.
\newblock URL \url{https://www.aclweb.org/anthology/P17-1099}.

\bibitem[Soltan et~al.(2022)Soltan, Ananthakrishnan, FitzGerald, Gupta, Hamza,
  Khan, Peris, Rawls, Rosenbaum, Rumshisky, et~al.]{alexatm}
Saleh Soltan, Shankar Ananthakrishnan, Jack FitzGerald, Rahul Gupta, Wael
  Hamza, Haidar Khan, Charith Peris, Stephen Rawls, Andy Rosenbaum, Anna
  Rumshisky, et~al.
\newblock Alexatm 20b: Few-shot learning using a large-scale multilingual
  seq2seq model.
\newblock \emph{arXiv preprint arXiv:2208.01448}, 2022.

\bibitem[Tay et~al.(2022)Tay, Dehghani, Tran, Garcia, Bahri, Schuster, Zheng,
  Houlsby, and Metzler]{unifying}
Yi~Tay, Mostafa Dehghani, Vinh~Q Tran, Xavier Garcia, Dara Bahri, Tal Schuster,
  Huaixiu~Steven Zheng, Neil Houlsby, and Donald Metzler.
\newblock Unifying language learning paradigms.
\newblock \emph{arXiv preprint arXiv:2205.05131}, 2022.

\bibitem[Vaswani et~al.(2017)Vaswani, Shazeer, Parmar, Uszkoreit, Jones, Gomez,
  Kaiser, and Polosukhin]{attention}
Ashish Vaswani, Noam Shazeer, Niki Parmar, Jakob Uszkoreit, Llion Jones,
  Aidan~N Gomez, {\L}ukasz Kaiser, and Illia Polosukhin.
\newblock Attention is all you need.
\newblock \emph{Advances in neural information processing systems}, 30, 2017.

\bibitem[Wang \& Cho(2019)Wang and Cho]{bertspeak}
Alex Wang and Kyunghyun Cho.
\newblock Bert has a mouth, and it must speak: Bert as a markov random field
  language model.
\newblock In \emph{Proceedings of the Workshop on Methods for Optimizing and
  Evaluating Neural Language Generation}, pp.\  30--36, 2019.

\bibitem[Wang \& Komatsuzaki(2021)Wang and Komatsuzaki]{gptj}
Ben Wang and Aran Komatsuzaki.
\newblock {GPT-J-6B: A 6 Billion Parameter Autoregressive Language Model}.
\newblock \url{https://github.com/kingoflolz/mesh-transformer-jax}, May 2021.

\bibitem[Wang et~al.(2022)Wang, Roberts, Hesslow, Scao, Chung, Beltagy, Launay,
  and Raffel]{bigsciencearchobjective}
Thomas Wang, Adam Roberts, Daniel Hesslow, Teven~Le Scao, Hyung~Won Chung,
  Iz~Beltagy, Julien Launay, and Colin Raffel.
\newblock What language model architecture and pretraining objective work best
  for zero-shot generalization?, 2022.

\bibitem[Wei et~al.(2021)Wei, Bosma, Zhao, Guu, Yu, Lester, Du, Dai, and
  Le]{flan}
Jason Wei, Maarten Bosma, Vincent~Y Zhao, Kelvin Guu, Adams~Wei Yu, Brian
  Lester, Nan Du, Andrew~M Dai, and Quoc~V Le.
\newblock Finetuned language models are zero-shot learners.
\newblock \emph{arXiv preprint arXiv:2109.01652}, 2021.

\bibitem[Wolf et~al.(2019)Wolf, Debut, Sanh, Chaumond, Delangue, Moi, Cistac,
  Rault, Louf, Funtowicz, and Brew]{transformers}
Thomas Wolf, Lysandre Debut, Victor Sanh, Julien Chaumond, Clement Delangue,
  Anthony Moi, Pierric Cistac, Tim Rault, R{\'{e}}mi Louf, Morgan Funtowicz,
  and Jamie Brew.
\newblock Huggingface's transformers: State-of-the-art natural language
  processing.
\newblock \emph{CoRR}, abs/1910.03771, 2019.
\newblock URL \url{http://arxiv.org/abs/1910.03771}.

\bibitem[Xue et~al.(2021)Xue, Constant, Roberts, Kale, Al-Rfou, Siddhant,
  Barua, and Raffel]{xue-etal-2021-mt5}
Linting Xue, Noah Constant, Adam Roberts, Mihir Kale, Rami Al-Rfou, Aditya
  Siddhant, Aditya Barua, and Colin Raffel.
\newblock m{T}5: A massively multilingual pre-trained text-to-text transformer.
\newblock In \emph{Proceedings of the 2021 Conference of the North American
  Chapter of the Association for Computational Linguistics: Human Language
  Technologies}, pp.\  483--498, Online, June 2021. Association for
  Computational Linguistics.
\newblock \doi{10.18653/v1/2021.naacl-main.41}.
\newblock URL \url{https://aclanthology.org/2021.naacl-main.41}.

\bibitem[Zhang et~al.(2022)Zhang, Roller, Goyal, Artetxe, Chen, Chen, Dewan,
  Diab, Li, Lin, et~al.]{opt}
Susan Zhang, Stephen Roller, Naman Goyal, Mikel Artetxe, Moya Chen, Shuohui
  Chen, Christopher Dewan, Mona Diab, Xian Li, Xi~Victoria Lin, et~al.
\newblock Opt: Open pre-trained transformer language models.
\newblock \emph{arXiv preprint arXiv:2205.01068}, 2022.

\bibitem[Zhang et~al.(2019)Zhang, Kishore, Wu, Weinberger, and
  Artzi]{bertscore}
Tianyi Zhang, Varsha Kishore, Felix Wu, Kilian~Q Weinberger, and Yoav Artzi.
\newblock Bertscore: Evaluating text generation with bert.
\newblock In \emph{International Conference on Learning Representations}, 2019.

\end{thebibliography}
